\documentclass[Afour,sageh,times]{sagej}

\usepackage{moreverb,url}
\usepackage{float, bm, caption, mathtools}
\usepackage[colorlinks,bookmarksopen,bookmarksnumbered,citecolor=red,urlcolor=red]{hyperref}

\newcommand\BibTeX{{\rmfamily B\kern-.05em \textsc{i\kern-.025em b}\kern-.08em
T\kern-.1667em\lower.7ex\hbox{E}\kern-.125emX}}

\def\volumeyear{2016}

\begin{document}
\setcounter{secnumdepth}{3}

\runninghead{Robot Control based on Motor Primitives --- A Comparison of Two Approaches}

\title{Robot Control based on Motor Primitives --- A Comparison \newline of Two Approaches}

\author{Moses C. Nah\affilnum{1}, Johannes Lachner\affilnum{1}, and Neville Hogan\affilnum{1,2}}

\affiliation{\affilnum{1}Massachusetts Institute of Technology, Department of Mechanical Engineering, USA.\\
\affilnum{2}Massachusetts Institute of Technology, Department of Brain and Cognitive Sciences, USA.}

\corrauth{Moses C. Nah, Massachusetts Institute of Technology, Department of Mechanical Engineering,
Cambridge, Massachusetts, USA.}

\email{mosesnah@mit.edu}

\begin{abstract}
Motor primitives are fundamental building blocks of a controller which enable dynamic robot behavior with minimal high-level intervention.
By treating motor primitives as basic ``modules,'' different modules can be sequenced or superimposed to generate a rich repertoire of motor behavior.
In robotics, two distinct approaches have been proposed: Dynamic Movement Primitives (DMPs) and Elementary Dynamic Actions (EDAs). 
While both approaches instantiate similar ideas, significant differences also exist. 
This paper attempts to clarify the distinction and provide a unifying view by delineating the similarities and differences between DMPs and EDAs. 
We provide eight robot control examples, including sequencing or superimposing movements, managing kinematic redundancy and singularity, obstacle avoidance, and managing physical interaction.
We show that the two approaches clearly diverge in their implementation. 
We also discuss how DMPs and EDAs might be combined to get the best of both approaches.
With this detailed comparison, we enable researchers to make informed decisions to select the most suitable approach for specific robot tasks and applications.
\end{abstract}


\maketitle

\section{Introduction}
\footnotetext{This manuscript has been submitted to the International Journal of Robotics Research for review.}
One of the major challenges of robotics is to generate complex motor behavior that can match that of humans \citep{billard2019trends}.
Numerous approaches have been developed to address this problem, including applied nonlinear control \citep{slotine1991applied, chung2009cooperative}, optimization-based approaches \citep{posa2014direct,kuindersma2016optimization}, and machine learning algorithms \citep{schulman2015trust,lillicrap2015continuous,haarnoja2018soft}.

Among these methods, several advances have been made based on ``motor primitives'' \citep{schaal1999imitation,ijspeert2013dynamical,hogan2012dynamic}.
The fundamental concept originates from human motor control research, where complex motor behavior of biological systems appears to be generated by a combination of fundamental building blocks known as motor primitives \citep{wolpert2001perspectives,mussa2000motor,bizzi2017acquisition,flash2005motor,hogan2012dynamic}.

The concept of control based on motor primitives dates back at least a century \citep{sherrington1906integrative}, with a number of subsequent experiments providing support for its existence in biological systems. 
Sherrington was one of the first to suggest ``reflex'' as a fundamental element of complex motor behavior \citep{sherrington1906integrative, elliott2001century}.
Sherrington proposed that reflexes can be treated as basic units of motor behavior that when chained together produce more complex movements \citep{clower1998early}.
First formalized by Bernstein \citep{bernstein1935problem,latash2021one}, ``synergies'' have also been suggested as a motor primitive to account for the simultaneous motion of multiple joints or activation of multiple muscles \citep{d2003combinations,giszter1993convergent,giszter2013motor}.
Discrete and rhythmic movements have also been suggested as two distinct classes of primitives \citep{schaal2004rhythmic, hogan2007rhythmic, hogan2012dynamic, park2017moving, viviani1995minimum, huh2015spectrum}.
Recently, there is growing evidence that ``stable postures'' may be considered to be a distinct class of motor primitives \citep{shadmehr2017distinct, jayasinghe2022neural}. 

Motor primitives have also been applied to robotics. 
Two distinct approaches exist: Dynamic Movement Primitives (DMPs) \citep{ijspeert2013dynamical,schaal2006dynamic} and Elementary Dynamic Actions (EDAs) \citep{hogan2012dynamic,hogan2013dynamic,hogan2017physical}.\footnote{The original name suggested by \cite{hogan2012dynamic} was ``Dynamic \emph{Motor} Primitives.'' However, to avoid confusion due to similarity to ``Dynamic \emph{Movement} Primitives,'' here we use the term ``Elementary Dynamic Actions'' (EDAs). The differences will be clarified in this paper. }
The key idea of these approaches is to formulate motor primitives as ``attractors'' \citep{hogan2012dynamic, ijspeert2013dynamical}. 
An attractor is a prominent feature of nonlinear dynamical systems, defined as a set of states towards which the system tends to evolve.
Its type ranges from relatively simple ones such as (stable) ``point attractors'' and (stable) ``limit cycles,'' to ``strange attractors'' \citep{strogatz2018nonlinear} such as the ``Lorenz attractor'' \citep{lorenz1963deterministic}, ``Rössler attractor'' \citep{rossler1976equation}, and others \citep{tam2008generation,sprott2014dynamical}. 

One of the key benefits of using motor primitives is that it enables highly dynamic behavior of the robot with minimal high-level intervention \citep{hogan2012dynamic}.
As a result, the complexity of the control problem can be significantly reduced.
For instance, by formulating discrete (respectively rhythmic) movement as a stable point attractor (respectively limit cycle), the problem of generating the movement reduces to learning the parameters of the corresponding attractor.    
Another important consequence is that it provides a modular structure of the controller. 
By treating motor primitives as basic ``modules,'' learning motor skills happens at the level of modules which provides adaptability and flexibility for robot control.

Since DMPs and EDAs stem from the theory of motor primitives, both approaches share the same philosophy.
Nevertheless, significant differences exist such that their implementations diverge. In the opinion of the authors, this has not yet been sufficiently emphasized.
An in-depth review that elucidates the similarities and differences between the two approaches may be beneficial to the robotics community.

In this paper, we provide a comprehensive review of motor primitives in robotics, focusing specifically on the two distinct approaches --- DMPs and EDAs.
We delineate the similarities and differences of both approaches by presenting eight extensively used robotic control examples (Section \ref{subsec:comparison}).\footnote{The code for the simulation examples is available in \url{https://github.com/mosesnah-shared/DMP-comparison}}
We show that:
\begin{itemize}
    \item Both approaches use motor primitives as basic building blocks to parameterize the controller (Section \ref{sec:theory_of_DMP_and_EDA}). DMPs consist of a canonical system, nonlinear forcing terms and transformation systems. EDAs consist of submovements, oscillations and mechanical impedances. 
    \item For torque-controlled robots, DMPs require an inverse dynamic model of the robot, whereas EDAs do not impose this requirement (Section \ref{subsec:inverse_dynamics_model}).
    \item With an inverse dynamics model, DMPs can achieve perfect tracking, both in task-space and joint-space (Section \ref{subsec:joint_space_trajectory_tracking}, \ref{subsec:task_space_traj_track_no_redund}). Imitation Learning enables DMPs to learn and track trajectories of arbitrary complexity (Section \ref{subsubsec:imitation_learning}). Online trajectory modulation of DMPs enables achieving additional control objectives such as obstacle avoidance (Section \ref{subsec:obstacle_avoidance}), thereby providing advantages over spline methods.
    For tracking control with EDAs, an additional method for calculating an appropriate virtual trajectory and mechanical impedance to which it is connected (Section \ref{subsubsec:NortonEN}) is required (Section \ref{subsec:dynamic_motor_primitives}, \ref{subsec:joint_space_trajectory_tracking}, \ref{subsec:task_space_traj_track_no_redund}).
    \item To control the end-effector of the robot, DMPs require additional control methods to manage kinematic singularity and kinematic redundancy (Section \ref{subsec:task_space_traj_track_no_redund}, \ref{subsec:managing_kin_redund}). In contrast, for EDAs, stability near (and even at) kinematic singularity can be ensured (Section \ref{subsec:task_space_traj_track_no_redund}). 
    Kinematic redundancy can be managed without solving the inverse kinematics (Section \ref{subsec:managing_kin_redund}). 
    \item Both approaches provide a modular framework for robot control. However, 
    the extent of modularity and its practical implications differ between the two approaches.
    A clear distinction appears when combining multiple movements. 
    For DMPs, discrete and rhythmic movements are represented by different DMPs (Section \ref{subsubsec:canonical_system}, \ref{subsubsec:nonlinear_forcing_term}). 
    Hence, the two different DMPs cannot be directly superimposed to generate a combination of discrete and rhythmic movements (Section \ref{subsec:superposition_of_discrete_rhythmic}). 
    Multiple discrete movements are generated by modifying the goal position of the previous movement (Section \ref{subsec:goal_changing}). 
    While the weights of the nonlinear forcing terms learned from Imitation Learning (Section \ref{subsubsec:imitation_learning}) can be reused, the weights of different DMPs cannot be simply combined.
    
    For EDAs, modularity exists both at the level of kinematics and mechanical impedances (Section \ref{subsubsec:NortonEN}). 
    For the former, multiple movements can be directly superimposed at the kinematic level, which provides a greater degree  of simplicity to generate a rich repertoire of movements, e.g., combining discrete and rhythmic movements (Section \ref{subsec:superposition_of_discrete_rhythmic}), or sequencing discrete movements (Section \ref{subsec:goal_changing}).
    For the latter, the learned mechanical impedances can be reused and combined by simple superposition. 
    Superposition of mechanical impedances enables a ``divide-and-conquer'' strategy, where complex tasks can be broken down into simpler sub-tasks.
    This modular property of mechanical impedances simplifies multiple control tasks, e.g., obstacle avoidance (Section \ref{subsec:obstacle_avoidance}) and managing kinematic redundancy (Section \ref{subsec:managing_kin_redund}).

    \item For DMPs, a low-gain PD controller is superimposed to manage uncertainty and physical contact (Section \ref{subsec:inverse_dynamics_model}). EDAs include mechanical impedance as a separate primitive to manage physical interaction (Section \ref{subsubsec:mechanical_impedances}). The dynamics of physical interaction can be controlled by modulating mechanical impedance (Section \ref{subsec:modulatingMechImpedances}). 
\end{itemize}
Lastly, we show how DMPs and EDAs might be integrated, thereby leveraging the best of both approaches.

\section{Theory}\label{sec:theory_of_DMP_and_EDA}
In this Section, we provide an overview of DMPs and EDAs. 
For simplicity, we consider a system with a single DOF.
A generalization to systems with multiple DOFs is presented in Section \ref{subsec:comparison}.

\subsection{Dynamic Movement Primitives}\label{subsec:dynamic_movement_primitives}
DMPs, introduced by \cite{schaal1999imitation,schaal2006dynamic,ijspeert2013dynamical}, consist of three classes of primitives: a canonical system (Section \ref{subsubsec:canonical_system}), a nonlinear forcing term (Section \ref{subsubsec:nonlinear_forcing_term}), and a transformation system (Section \ref{subsubsec:transformation_system}). 
To generate discrete and rhythmic movements, two distinct definitions exist for canonical system and nonlinear forcing term.
For clarification, labels ``Discrete'' and ``Rhythmic'' are added next to the equation.

\subsubsection{Canonical System}\label{subsubsec:canonical_system}
A canonical system $s:\mathbb{R}_{\ge 0}\rightarrow \mathbb{R}_{\ge 0}$ is a scalar variable governed by a first-order differential equation:
\begin{equation}\label{eq:DMP_canonical_system}
    \begin{aligned}
        \tau \dot{s}(t) = 
        \begin{cases}
          -\alpha_s s(t) & \text{ \footnotesize Discrete } \\
                    1 & \text{ \footnotesize Rhythmic}
        \end{cases}
    \end{aligned}
\end{equation}
In these equations, $\alpha_s$ is a positive constant, $t\in\mathbb{R}_{\ge 0}$ is time and $\tau>0$ is a time constant.

For discrete movements, the canonical system is exponentially convergent to 0 with a closed-form solution $s(t)=\exp(-\alpha_st/\tau)s(0)$.
For rhythmic movements, the canonical system is a linear function of time $s(t)=t/\tau$, but the modulo-$2\pi$ operation is applied to ensure $s\in[0,2\pi)$.

\subsubsection{Nonlinear Forcing Term}\label{subsubsec:nonlinear_forcing_term}
A nonlinear forcing term $f:\mathbb{R}_{\ge 0}\rightarrow \mathbb{R}$, which takes the canonical system $s(t)$ as the function argument, is defined by:
\begin{equation}\label{eq:DMP_nonlinear_force}
    \begin{aligned}
        f(s(t)) = 
        \begin{cases}
           \frac{\sum_{i=1}^{N}w_i\phi_i(s(t))}{\sum_{i=1}^{N}\phi_i(s(t))} s(t)(g-y_0) &   \text{ \footnotesize Discrete } \\
                    \frac{\sum_{i=1}^{N}w_i\phi_i(s(t))}{\sum_{i=1}^{N}\phi_i(s(t))} r & \text{ \footnotesize Rhythmic }
        \end{cases}
    \end{aligned}
\end{equation}
In Equation \eqref{eq:DMP_nonlinear_force},
$N$ is the number of basis functions; $w_i$ is the weight of the $i$-th basis function; $y_0$ and $g$ are the initial and final positions of the discrete movement, respectively; $r$ is the amplitude of the nonlinear forcing term for rhythmic movements and $\phi_i:\mathbb{R}_{\ge 0}\rightarrow \mathbb{R}$ is the $i$-th basis function of the nonlinear forcing term:
\begin{equation}\label{eq:DMP_basis_functions}
    \begin{aligned}
        \phi_i(s(t)) = 
        \begin{cases}
           \exp\big\{ -h_i(s(t)-c_i)^2 \big\} & \text{\footnotesize Discrete } \\  
           \exp\big\{  h_i (\cos(s(t)-c_i)-1) \big\} & \text{\footnotesize Rhythmic }
        \end{cases}
    \end{aligned}
\end{equation}
In Equation \eqref{eq:DMP_basis_functions}, the basis functions for discrete and rhythmic movements are Gaussian functions and von Mises functions, respectively \citep{ijspeert2013dynamical}; $c_i$ is the center of the $i$-th basis function; $h_i$ is a positive constant that determines the width of the $i$-th basis function.

\subsubsection{Transformation System}\label{subsubsec:transformation_system}
The nonlinear forcing term $f$, with canonical system $s$ as its function variable, is used as an input to the transformation system to generate trajectories with arbitrary complexity.
In detail, a transformation system is a second-order differential equation, which is equivalent to a linear mass-spring-damper model with a nonlinear force input $f(s(t))$:
\begin{equation}\label{eq:DMP_transformation}
    \begin{aligned}
        \tau \dot{y}(t) &= z(t) \\
        \tau \dot{z}(t) &= \alpha_z \{ \beta_z (g-y(t)) -z(t)\} + f(s(t))
    \end{aligned}    
\end{equation}
In these equations, $\alpha_z$ and $\beta_z$ are positive constants; $y(t)$ and $z(t)$ are state variables which correspond to position and (time-scaled) velocity of the transformation system, respectively; $\tau$ is the time constant used for the canonical system (Equation \eqref{eq:DMP_canonical_system}); for discrete movement, $g$ is identical to Equation \eqref{eq:DMP_nonlinear_force}; 
for rhythmic movement, $g$ is chosen to be the average of the rhythmic, repetitive movement \citep{ijspeert2013dynamical}.

While any positive values of $\alpha_z$ and $\beta_z$ can be used, usually, the values of $\alpha_z$ and $\beta_z$ are chosen such that if $f(s(t))$ is zero, the transformation system is critically damped (i.e., has repeated eigenvalues) for $\tau=1$ (i.e., $\beta_z=\alpha_z/4$) \citep{ijspeert2013dynamical}.

Using the canonical system $s$ avoids the explicit time-dependency of the transformation system and results in an autonomous system \citep{ijspeert2013dynamical}. 
Moreover, while both discrete and rhythmic movements are generated with the same transformation system, different choices of canonical systems $s$ and nonlinear forcing terms $f$ are made to produce those movements.
Hence, to produce a combination of discrete and rhythmic movements, both discrete and rhythmic DMPs must be constructed (Section \ref{subsec:superposition_of_discrete_rhythmic}).

\subsubsection{Imitation Learning}\label{subsubsec:imitation_learning}
One prominent application of DMPs is ``Imitation Learning,'' also called ``Learning from Demonstration'' \citep{schaal1999imitation, ijspeert2001trajectory, ijspeert2002movement, ijspeert2013dynamical}.
Let $y_{des}(t)$ be the desired trajectory that the robot aims to learn (or imitate). 
Then, $y(t)=y_{des}(t)$ can be achieved by using the following nonlinear forcing term (Equation \eqref{eq:DMP_transformation}):
\[
    f_{target}(s(t)) = \tau^2\ddot{y}_{des}(t) + \alpha_z \tau\dot{y}_{des}(t) + \alpha_z\beta_z(y_{des}(t)-g)
\]
If the analytic solution of $y_{des}(t)$ and its derivatives are not known, Imitation Learning generates the whole continuous trajectory of $y_{des}(t)$ from $P$ sample points, $(y_{des}(t_i),$ $\dot{y}_{des}(t_i)$, $\ddot{y}_{des}(t_i))$ for $i\in[1,2,\cdots, P]$, and the $P$ sample points are used to find the best-fit weights of $f(s(t))$ (Equation \eqref{eq:DMP_basis_functions}) that matches $f_{target}(s(t))$.

The best-fit weight $w_i^*$ of the $i$-th basis function is calculated using Locally Weighted Regression \citep{atkeson1997locally,schaal1998constructive,ijspeert2013dynamical}:
\begin{equation}\label{eq:DMP_batch_regression}
    w_i^* = \frac{\mathbf{a}^{\text{T}} \Phi_i\mathbf{f}_{target}}{\mathbf{a}^{\text{T}} \Phi_i \mathbf{a}}
\end{equation}
where:
\begin{align*}
    \mathbf{a} &= 
    \begin{bmatrix}
        a_1 \\
        a_2 \\
        \vdots \\
        a_P \\
    \end{bmatrix}
    \quad 
    \mathbf{f}_{target} = 
    \begin{bmatrix}
        f_{target,1} \\
        f_{target,2} \\
        \vdots \\
        f_{target,P} \\
    \end{bmatrix}    
    \\
    \Phi_i &= 
    \begin{bmatrix}
        \phi_i( s( t_1 ) ) &  & & 0 \\
         &  \phi_i( s( t_2 ) ) & &  \\
         & & \ddots &   \\
         0 & & & \phi_i( s( t_P ) )\\
    \end{bmatrix} 
\end{align*}        
The elements of $\mathbf{a}$ and $\mathbf{f}_{target}$ are:
\begin{align*}
    a_i\equiv a(t_i) &= 
    \begin{cases}
       s(t_i)(g-y_0) & \text{ \footnotesize Discrete } \\ 
       r             & \text{ \footnotesize Rhythmic }
    \end{cases} \\
    f_{target,i} &= \tau^2\ddot{y}_{des}(t_i) + \alpha_z \tau\dot{y}_{des}(t_i) + \alpha_z\beta_z(y_{des}(t_i)-g) 
\end{align*}
Along with Locally Weighted Regression, one can also use linear least square regression to find the best fit weights \citep{ude2014orientation,saveriano2019merging}.

For Imitation Learning of discrete movements, the goal $g$ and the initial position $y_0$ are set as $g=y_{des}(t_P)$ and $y_0=y_{des}(t_1)$, respectively; $\tau$ is chosen to be the duration of the discrete movement.
For Imitation Learning of rhythmic movements, goal $g$ is the midpoint of the minimum and maximum values of $y_{des}(t_1), y_{des}(t_2), \cdots, y_{des}(t_P)$; $\tau$ is chosen to be the period of the demonstrated movement divided by $2\pi$, hence the period of the rhythmic movement must be derived first \citep{ijspeert2013dynamical}.

With these best-fit weights $w_i^*$, the best-fit nonlinear forcing term $f^*(s(t))$ is derived and used as the input to the transformation system to generate $y_{des}(t)$, $\dot{y}_{des}(t)$, $\ddot{y}_{des}(t)$.

One might wonder why spline methods are not used to derive $y_{des}(t)$, $\dot{y}_{des}(t)$, $\ddot{y}_{des}(t)$ from the $P$ sample points \citep{wada2004via}.
Splines are effective for smooth trajectory generation and are widely used in industrial robotics. However, spline methods do not allow online trajectory modulation of DMPs \citep{ijspeert2013dynamical} which is crucial to achieve multiple control tasks, e.g., obstacle avoidance (Section \ref{subsec:obstacle_avoidance}) and sequencing discrete movements (Section \ref{subsec:goal_changing}).
Compared to spline methods, Imitation Learning provides modularity for robot control. 
Once the best-fit weights are learned, these weights can be saved as a learned module, which can be reused to generate the learned trajectory (Section \ref{subsec:task_space_traj_track_no_redund}, \ref{subsec:obstacle_avoidance}). 
Finally, DMPs provide favorable invariance properties. 
Once the best-fit weights of the demonstrated trajectory are learned, the learned movement can be scaled in time (i.e., temporal invariance) and space (i.e., spatial invariance) without changing the qualitative property of the movement \citep{ijspeert2013dynamical}.

If the $P$ sample points of a demonstrated trajectory are provided, the best-fit weights which produce that demonstrated trajectory can be calculated with matrix algebra, referred to as ``one-shot learning'' (Equation \eqref{eq:DMP_batch_regression}).
This process is called ``batch regression'' \citep{ijspeert2013dynamical}, since all $P$ data points should be collected to calculate the $N$ weights. 
The batch regression method assumes a predefined number of basis functions $N$ and parameters $c_i$ and $h_i$.
While these parameters can be defined manually, the Locally Weighted Projection Regression method \citep{vijayakumar2000locally} can identify the necessary number of basis functions $N$, the center locations $c_i$, and width parameters $h_i$ using nonparametric regression techniques.

Note that Imitation Learning for joint trajectories is easily scalable to high DOF systems \citep{atkeson2000using,ijspeert2002movement}.
For an $n$-DOF system, one can construct $n$ transformation systems, each representing a joint trajectory. 
The $n$ transformation systems are synchronized with a single canonical system \citep{ijspeert2013dynamical}.
With the Imitation Learning method, learning the best-fit weights is computationally efficient as it involves simple matrix algebra.

\subsection{Elementary Dynamic Actions}\label{subsec:dynamic_motor_primitives}
\begin{figure}
    \centering
    \includegraphics[trim={0.0cm 0.5cm 0.0cm 0.0cm}, width=0.90\columnwidth, clip, page=1]{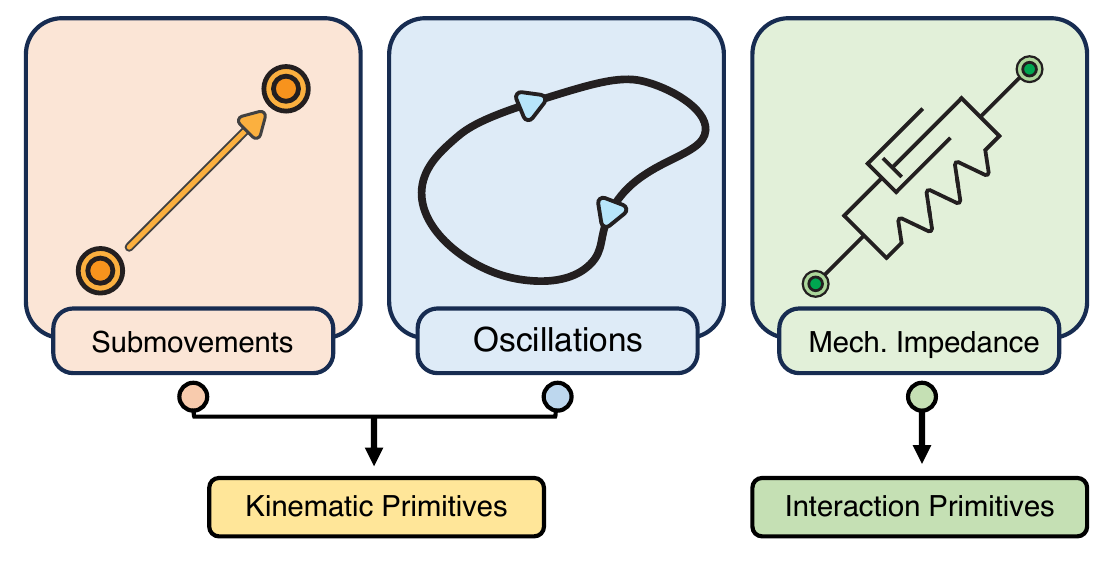}
    \caption{The three primitives of Elementary Dynamic Actions (EDAs). Submovements and oscillations correspond to kinematic primitives and mechanical impedances manage physical interaction.}
    \label{fig:edas}        
\end{figure}

EDAs, introduced by \cite{hogan2012dynamic,hogan2013dynamic} consist of at least three distinct classes of primitives: submovements (Section \ref{subsubsec:submovement}) and oscillations (Section \ref{subsubsec:oscillation}) as kinematic primitives, and mechanical impedances as interaction primitives (Section \ref{subsubsec:mechanical_impedances}) (Figure \ref{fig:edas}).

\subsubsection{Submovements}\label{subsubsec:submovement}
A submovement $x_0:\mathbb{R}_{\ge 0}\rightarrow \mathbb{R}$ is a smooth trajectory in which its time derivative is a unimodal function, i.e., has a single peak value:
\begin{equation*}
    \dot{x}_0(t) = v \ \hat{\sigma}(t)
\end{equation*}
In this equation, $\hat{\sigma}:\mathbb{R}_{\ge 0}\rightarrow [0,1]$ denotes a smooth unimodal basis function with peak value 1; 
$v\in\mathbb{R}$ is the velocity amplitude of the submovement. 

Submovements model discrete motions, and therefore $\hat{\sigma}(t)$ has a finite support, i.e., there exists $T>0$ such that $\hat{\sigma}(t)=0$ for $t\ge T$. 
\citep{hogan2007rhythmic}.
The shape of $\hat{\sigma}(t)$ can either be symmetric or not. 

\subsubsection{Oscillations}\label{subsubsec:oscillation}
An oscillation $x_0:\mathbb{R}_{\ge 0}\rightarrow \mathbb{R}$ is a smooth non-zero trajectory which is a periodic function:
\begin{equation*}
    \forall t >0: ~~ \exists T>0: ~~ x_0(t) = x_0(t+T)
\end{equation*}
Note that this definition of oscillation can be too strict and the definition of oscillation can be expanded to almost-periodic functions \citep{hogan2012dynamic}.   
For our purposes, it is sufficient to think of an oscillation as a periodic function.
Compared to submovements, oscillations model rhythmic and repetitive motions.

\subsubsection{Mechanical Impedances}\label{subsubsec:mechanical_impedances}
Mechanical impedance $Z$ is an operator which maps (generalized) displacement $\Delta x(t) \in \mathbb{R}$ to (generalized) force $F(t)\in\mathbb{R}$ \citep{hogan2017physical, hogan2018impedance}:
\begin{equation}\label{eq:mechanical_impedances}
    Z: \Delta x(t) \longrightarrow F(t)
\end{equation}
In detail, $\Delta x(t)$ is the displacement from an actual trajectory of (generalized) position $x(t)$ and a virtual trajectory $x_0(t)$ to which the mechanical impedance is connected, i.e., $\Delta x(t)=x_0(t)-x(t)$.
Loosely speaking, mechanical impedance is a generalization of stiffness to encompass nonlinear dynamic behavior \citep{hogan2017physical}.

The impedance operator of Equation \eqref{eq:mechanical_impedances} can denote both a map from joint displacement to torque, or a map from end-effector (generalized) displacement to (generalized) force. The former impedance operator is often referred to as ``joint-space impedance,'' and the latter is often referred to as ``task-space impedance.''
For task-space impedance, both translational \citep{hogan1985impedance} and rotational displacement \citep{caccavale1998quaternion} of the end-effector can be considered separately.

Along with the kinematic primitives, i.e., submovements and oscillations, EDAs include mechanical impedance as a distinct primitive to manage physical interaction \citep{hogan2017physical, hogan2018impedance,dietrich2022control,hogan2022contact}.
The dynamics of physical interaction can be controlled by modulating mechanical impedance. 
For instance, tactile exploration and manipulation of fragile objects should evoke the use of low stiffness, while tasks such as drilling a hole on a surface requires high stiffness for object stabilization \citep{hogan2018impedance}.

Under the assumption that the environment is an admittance, mechanical impedances can be linearly superimposed even though each mechanical impedance is a nonlinear operator. This is the superposition principle of mechanical impedances \citep{hogan1985impedance,hogan2017physical}:
\begin{equation}\label{eq:superposition_of_mechanical_impedances}
    Z = \sum Z_i  
\end{equation}
This principle provides a modular framework for robot control that can simplify multiple control tasks, e.g., obstacle avoidance (Section \ref{subsec:obstacle_avoidance}) or managing kinematic redundancy (Section \ref{subsec:managing_kin_redund}).

Note that the impedance operators of Equation \eqref{eq:superposition_of_mechanical_impedances} can include transformation maps. 
For instance, to superimpose a joint-space impedance and a task-space impedance at the torque level, the task-space impedance is multiplied by a Jacobian transpose to map from end-effector (generalized) force to joint torques (Section \ref{subsec:motor_primitives_inverse_dynamics_model}). 

The choice of mechanical impedance decides whether the virtual trajectory $x_0(t)$ is an attractor or repeller. 
The former can be used to produce discrete point-to-point movements (Section \ref{subsec:joint_space_trajectory_tracking}, \ref{subsec:task_space_traj_track_no_redund}), whereas the latter can be exploited for obstacle avoidance (Section \ref{subsec:obstacle_avoidance}) \citep{andrews1983impedance,newman1987high,khatib1986potential,hjorth2020energy}.

\subsubsection{Norton Equivalent Network Model}\label{subsubsec:NortonEN}
\begin{figure}[H]
    \centering
    \includegraphics[trim={0.0cm 0.0cm 0.0cm 0.0cm}, width=0.96\columnwidth, clip, page=1]{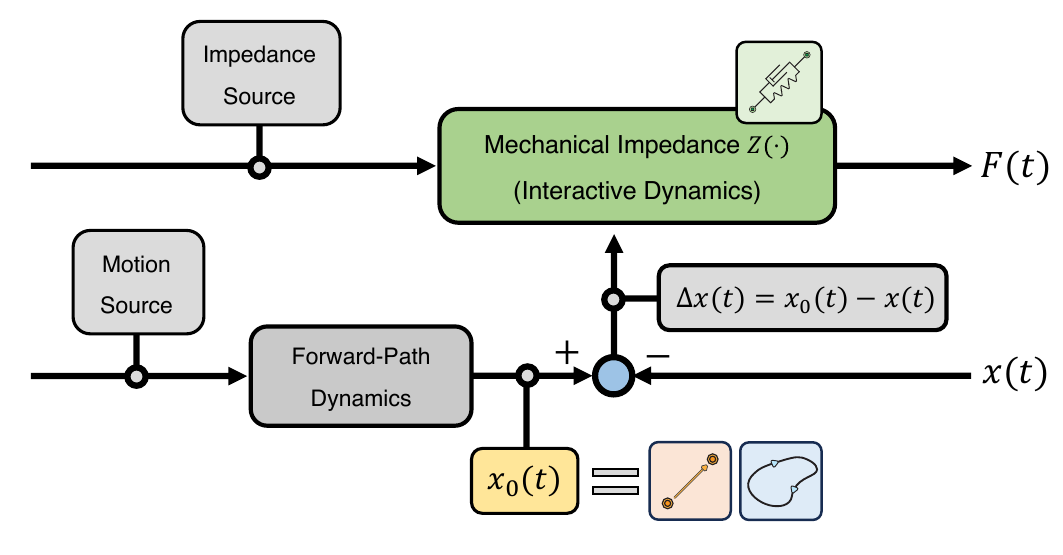}
    \caption{The three Elementary Dynamic Actions (EDAs) combined using a Norton equivalent network model. The virtual trajectory $x_0(t)$ (yellow box)  consists of submovements (orange box) and/or oscillations (blue box), and mechanical impedance $Z$ (green box) regulates the interactive dynamics.
     }
    \label{fig:edas_w_Norton_Network}        
\end{figure}

The three distinct classes of EDAs --- submovements, oscillations, and mechanical impedances --- may be combined using a Norton equivalent network model \citep{hogan2017physical}, which provides an effective framework to relate these classes of primitives (Figure \ref{fig:edas_w_Norton_Network}). 

In detail, the forward-path dynamics (Figure \ref{fig:edas_w_Norton_Network}) specifies the virtual trajectory $x_0(t)$, which consists of submovements and/or oscillations. 
The interactive dynamics, which consists of mechanical impedances $Z$, determines the generalized force output $F(t)$ with the generalized displacement input $\Delta x(t)$. 
Since the force output $F(t)$ is determined by the choice of virtual trajectory $x_0(t)$ and mechanical impedance $Z$, a key objective of EDAs is to find appropriate choices of $x_0(t)$ and $Z$ to produce the desired robot behavior. 

Note that submovements and/or oscillations can be directly combined at the level of the virtual trajectory $x_0(t)$, which thereby provides modularity at the kinematic level. 
As with the superposition principle of mechanical impedances (Section \ref{subsubsec:mechanical_impedances}), the modular property at the kinematic level provides several benefits for multiple control tasks, e.g., combining discrete and rhythmic movements (Section \ref{subsec:superposition_of_discrete_rhythmic}), sequencing discrete movements (Section \ref{subsec:goal_changing}).

As shown in Figure \ref{fig:edas_w_Norton_Network}, EDAs neither control $x(t)$ (i.e., position control) nor $F(t)$ (i.e., force/torque control) directly. 
Hence, EDAs are fundamentally different from position control \citep{tanner1981industrial,arimoto1984stability}, force/torque control \citep{whitney1977force}, or hybrid position/force control methods \citep{raibert1981hybrid}. 
This is one of the key ideas of impedance control.
This is also the reason why we have chosen the terminology ``virtual trajectory'' for $x_0(t)$ instead of a ``desired'' or a ``reference trajectory.''  Compared to tracking control methods which aim to follow a reference trajectory, $x_0(t)$ of EDAs is simply a virtual trajectory to which the impedances are connected. 

This property of EDAs has several benefits for robot control with physical interaction. 
Compared to $x(t)$ and $F(t)$ that depend on the environment or object with which the robot interacts, the virtual trajectory $x_0(t)$ and impedance operator $Z$ can be modulated ``independently,'' i.e., regardless of the environment or the manipulated object \citep{hogan2018impedance}.
For instance, force/torque control cannot be used for free-space motions and position control cannot be used in contact with a kinematically constrained environment. 
EDAs can be used for both cases since neither $x(t)$ nor $F(t)$ is controlled directly. 

Note that the Norton equivalent network model separates forward-path dynamics (virtual trajectory $x_0(t)$) from interactive dynamics (mechanical impedance $Z$).
Hence, parallel optimization of $x_0(t)$ and $Z$ can be conducted.
This has computational advantages for real-time control of robots with many DOF \citep{lachner_shaping_2022}.

\section{Comparison of the Two Approaches}\label{subsec:comparison}
In this Section, a detailed comparison between DMPs and EDAs is presented. 
To emphasize the similarities and differences between the two approaches, multiple simulation examples using the MuJoCo physics engine (Version 1.50) \citep{todorov2012mujoco} are presented. The code is available at \url{https://github.com/mosesnah-shared/DMP-comparison}.
A list of examples, ordered in progressive complexity, is shown below:
\begin{itemize}
    \item A goal-directed discrete movement in joint-space (Section \ref{subsec:joint_space_trajectory_tracking}).
    \item A goal-directed discrete movement in task-space (Section \ref{subsec:task_space_traj_track_no_redund}). 
    \item A goal-directed discrete movement in task-space, with unexpected physical contact (Section \ref{subsec:modulatingMechImpedances}).
    \item A goal-directed discrete movement in task-space, including obstacle avoidance (Section \ref{subsec:obstacle_avoidance}).
    \item Rhythmic movement, both in joint-space and task-space (Section \ref{subsec:rhythmic_movement}).
    \item Combination of discrete and rhythmic movements, both in joint-space and task-space (Section \ref{subsec:superposition_of_discrete_rhythmic}). 
    \item A sequence of discrete movements in task-space (Section \ref{subsec:goal_changing}). 
    \item A single (or sequence of) discrete movement(s) in task-space, while managing kinematic redundancy (Section \ref{subsec:managing_kin_redund}).
\end{itemize}
Some of the examples reproduce human-subject experiments in motor control research, e.g., \cite{burdet2001central} for Section \ref{subsec:task_space_traj_track_no_redund}, \cite{flash1991arm} for Section \ref{subsec:goal_changing}.

While in general position (or motion) control can be used to encode motor primitives, we will focus on the control of torque-actuated robots.
Position control would create challenges that restrict the set of tasks that can be achieved \citep{hogan2022contact}.
For instance, one of the challenges is the kinematic transformation from task-space coordinate to the robot’s joint configuration, which complicates control in task-space.
Another challenge is for tasks involving contact and physical interaction, which requires some level of compliance of the robotic manipulator.
For the eight simulation examples, we highlight the challenges when using position-actuated robots, e.g., managing contact and physical interaction (Section \ref{subsec:modulatingMechImpedances}), managing kinematic redundancy (Section \ref{subsec:managing_kin_redund}).
We show that torque-actuated robots can address these tasks without imposing such challenges.
Further discussion is deferred to Section \ref{subsubsec:requirement_of_inverse_kinematics}.

Given a torque-actuated open-chain $n$-DOF robot manipulator, its dynamics is governed by the following differential equation \citep{murray1994mathematical}:
\begin{equation}\label{eq:robot_manipulator_equation}
    \mathbf{M}(\mathbf{q})\ddot{\mathbf{q}} + \mathbf{C}(\mathbf{q}, \dot{\mathbf{q}})\dot{\mathbf{q}} + \mathbf{G}(\mathbf{q}) = \bm{\tau}_{in}(t) + \bm{\tau}_{ext}(t)
\end{equation}
In this equation, $\mathbf{q}\equiv \mathbf{q}(t)\in\mathbb{R}^n$ is the joint trajectory of the robot; $\mathbf{M}(\mathbf{q})\in\mathbb{R}^{n\times n}$ and $\mathbf{C}(\mathbf{q}, \dot{\mathbf{q}})\in\mathbb{R}^{n\times n}$ are the mass and centrifugal/Coriolis matrices, respectively; $\mathbf{G}(\mathbf{q})\in\mathbb{R}^{n}$ is the vector arising through gravitational potential energy;  $\bm{\tau}_{in}(t)\in\mathbb{R}^{n}$ is the torque input; $\bm{\tau}_{ext}(t)\in\mathbb{R}^{n}$ is the resultant effect of external forces expressed as torque.
For torque-controlled robots, the goal is to determine the torque input $\bm{\tau}_{in}(t)$ which produces desired robot behavior. 
For brevity and to avoid clutter, we often omit argument $t$.

For the presented examples, the orientation of the end-effector is not considered, since the control of end-effector's position suffices to illustrate the differences between the two approaches.
The details of implementing controllers for orientation can be found in references such as \cite{pastor2011online,abu2015adaptation} for DMPs and \cite{lachner2022geometric} for EDAs. 

In this paper, we use $\mathbf{p}\equiv\mathbf{p}(t)\in\mathbb{R}^{3}$ to denote the 3D Cartesian position of the end-effector and $\mathbf{h}$ to represent the Forward Kinematic Map of the robot, i.e., $\mathbf{p}=\mathbf{h(q)}$.
Except for tasks with physical contact (e.g., Section \ref{subsec:modulatingMechImpedances}), we assume $\bm{\tau}_{ext}(t)=\mathbf{0}$.
Finally, we assume that gravitational force $\mathbf{G}(\mathbf{q})$ is compensated by the controller and can be neglected (Equation \eqref{eq:robot_manipulator_equation}). 

\subsection{The Existence of an Inverse Dynamics Model}\label{subsec:inverse_dynamics_model}
We first show that for torque-actuated robots, DMPs require an inverse dynamics model, whereas EDAs do not.

\subsubsection{Dynamic Movement Primitives}
For DMPs, the transformation system (Equation \eqref{eq:DMP_transformation}) represents kinematic relations. 
Hence for a torque-actuated robot, the approach requires an inverse dynamics model, which determines the feedforward joint torques $\bm{\tau}_{in}(t)$ to generate the desired joint trajectory specified by $\ddot{\mathbf{q}}(t)$, $\dot{\mathbf{q}}(t)$, $\mathbf{q}(t)$ \citep{ijspeert2013dynamical}.
The inverse dynamics model requires an exact model of the robot manipulator, i.e., exact values of $\mathbf{M}(\mathbf{q})$ and $\mathbf{C}(\mathbf{q}, \dot{\mathbf{q}})$.

The requirement of an inverse dynamics model also implies that the DMP approach is in principle, a non-reactive feedforward control approach. 
To be robust against uncertainty or unexpected physical contact, a low-gain feedback control (e.g., PD control) is added to the feedforward inverse dynamics controller \citep{schaal2007dynamics,pastor2013dynamic} (Section \ref{subsec:modulatingMechImpedances}).
Moreover, for control in task-space where the maps from $\ddot{\mathbf{p}}(t)$, $\dot{\mathbf{p}}(t)$, $\mathbf{p}(t)$ to $\ddot{\mathbf{q}}(t)$, $\dot{\mathbf{q}}(t)$, $\mathbf{q}(t)$ are not always well defined, additional control methods must be employed.
For instance, to solve inverse kinematics, methods presented by \cite{whitney1969resolved, liegeois1977automatic, maciejewski1988numerical} are used \citep{park2008movement, nakanishi2008operational, pastor2009learning} (Section \ref{subsubsec:managing_kinematic_redundancy}).
For tracking control in task-space, feedback control methods such as sliding-mode control \citep{slotine1991applied,nakanishi2008operational}) are employed (Section \ref{subsubsec:managing_kinematic_redundancy}).

\subsubsection{Elementary Dynamic Actions}\label{subsec:motor_primitives_inverse_dynamics_model}
For EDAs, an inverse dynamics model is not required for a torque-actuated robot.
Hence, an exact model of the robot, i.e., exact computation of $\mathbf{M}(\mathbf{q})$ and $\mathbf{C}(\mathbf{q}, \dot{\mathbf{q}})$ is not essential, unless advanced applications such as inertia shaping \citep{khatib1995inertial,dietrich2019hierarchical} are desired.
Instead, the input torque command $\bm{\tau}_{in}(t)$ is determined by superimposing mechanical impedances (Section \ref{subsubsec:mechanical_impedances}):
\begin{equation*}
    \bm{\tau}_{in}(t) = \sum_{i}\mathbf{J}(\mathbf{q}(t))^{\text{T}}\mathbf{Z}_{p,i}(\Delta \mathbf{p}(t), t) + \sum_{i}\mathbf{Z}_{q,i}(\Delta \mathbf{q}(t), t)
\end{equation*}
In this equation, $\mathbf{J}(\mathbf{q}(t))\equiv \mathbf{J}(\mathbf{q})$ is the Jacobian matrix \citep{siciliano2008springer}; $\mathbf{Z}_p$ and $\mathbf{Z}_q$ denote task-space and joint-space impedances, respectively. 

Compared to DMPs, the EDA approach is in principle, a reactive feedback control method. 
Instead of an inverse dynamics model, measurements $\mathbf{q}(t)$ and a Forward Kinematics Map of the robot are required.
Given these values, one can react to the environment by modulating impedance and/or the virtual trajectories to regulate the dynamics of physical interaction (Section \ref{subsec:modulatingMechImpedances}).
Moreover, with appropriate choices of $\mathbf{Z}_p$, $\mathbf{Z}_q$ and $\mathbf{p}_0(t)$, $\mathbf{q}_0(t)$, the controller preserves passivity, which thereby provides robustness against uncertainty and external disturbances (Section \ref{subsec:modulatingMechImpedances}).

\subsection{A Goal-directed Discrete Movement in Joint-space}\label{subsec:joint_space_trajectory_tracking}
We consider designing a controller to generate a goal-directed discrete movement planned in joint-space coordinates. 

\subsubsection{Dynamic Movement Primitives}\label{subsubsec:joint_space_discrete_DMPs}
The movement of each joint is represented by a transformation system.
The $n$ transformation systems are synchronized by using a single discrete canonical system as the input to the $n$ nonlinear forcing terms \citep{ijspeert2013dynamical} (Section \ref{subsubsec:imitation_learning}):
\begin{align}\label{eq:example_movement_primitives_nDOF}
    \begin{aligned}
    \tau^2\ddot{\mathbf{q}}(t) &= -\alpha_z\beta_z \{\mathbf{q}(t)-\mathbf{g}\} - \alpha_z\tau\dot{\mathbf{q}}(t) + \mathbf{f}(s(t))\\
          \tau\dot{s}(t) &= -\alpha_s s(t)
    \end{aligned}
\end{align}
Note that different values of $\alpha_z$ and $\beta_z$ can be used for each transformation system. Nevertheless, it is sufficient to use identical values of $\alpha_z$ and $\beta_z$ for $n$ transformation systems.

Without using Imitation Learning, i.e., $\mathbf{f}(s(t))=\mathbf{0}$, the goal-directed discrete movement can be generated by setting $\mathbf{g}$ as the goal location. 
As a result, $\mathbf{q}(t)$ generates a motion of a stable second-order linear system which converges to $\mathbf{g}$.

In case we want to generate a goal-directed discrete movement that also follows a specific joint trajectory, Imitation Learning can be used. 
Let $\mathbf{q}_{des}(t)$ be the desired discrete joint-trajectory with duration $T$, such that $\mathbf{q}_{des}(T)=\mathbf{g}$ and $\tau=T$. 
The best-fit force $\mathbf{f}^*(s(t))$ is calculated and used as the input to $n$ transformation systems to produce $\ddot{\mathbf{q}}_{des}(t)$ (Equation \eqref{eq:example_movement_primitives_nDOF}).
With the initial conditions $\mathbf{q}_{des}(0), \dot{\mathbf{q}}_{des}(0)$, the trajectories of $\mathbf{q}_{des}(t)$, $\dot{\mathbf{q}}_{des}(t)$ are calculated using numerical integration. 
The calculated $\mathbf{q}_{des}(t)$, $\dot{\mathbf{q}}_{des}(t)$, $\ddot{\mathbf{q}}_{des}(t)$ are the input to the inverse dynamics model to generate the necessary torque input $\bm{\tau}_{in}(t)$.

\subsubsection{Elementary Dynamic Actions}\label{subsubsec:joint_space_discrete_EDAs}
To generate a goal-directed discrete movement planned in joint-space coordinates, we construct the following controller:
\begin{equation}\label{eq:joint_space_impedance_controller}
   \bm{\tau}_{in}(t) = \mathbf{K}_q \{\mathbf{q}_{0}(t) -\mathbf{q} \} + \mathbf{B}_q \{ \dot{\mathbf{q}}_{0}(t)-\dot{\mathbf{q}} \} 
\end{equation}
In this equation, $\mathbf{K}_q, \mathbf{B}_q\in\mathbb{R}^{n\times n}$ are symmetric positive definite matrices which correspond to the joint stiffness and damping, respectively.
This controller is a first-order joint-space impedance controller \citep{nah2020dynamic,nah2021manipulating}.

To generate a goal directed discrete movement, $\mathbf{q}_0(t)$ is chosen to be a submovement which ends at goal location $\mathbf{g}$, i.e., if the duration of submovement is $T$, then $\mathbf{q}_0(T)=\mathbf{g}$.
For constant positive definite $\mathbf{K}_q, \mathbf{B}_q$ matrices, $\mathbf{q}(t)$ asymptotically converges to $\mathbf{g}$ \citep{takegaki1981new,slotine1991applied}.

\subsubsection{Simulation Example}\label{subsubsec:joint_space_discrete_simulation_example}
Consider a 2-DOF planar robot model, where each link consists of a single uniform slender bar with mass and length of 1kg and 1m, respectively. Let the initial joint configuration be $\mathbf{q}(0)=\mathbf{q}_{i}\in\mathbb{R}^2$.
The code script for this simulation is \texttt{main\_joint\_discrete.py}.

For $\mathbf{q}_{des}(t)$ of DMPs and $\mathbf{q}_{0}(t)$ of EDAs, both trajectories were set to be a minimum-jerk trajectory \citep{flash1985coordination, hogan1987moving}:
\begin{align}\label{eq:minimum_jerk_trajectory}
\begin{split}
    &\mathbf{q}_{des}(t), \mathbf{q}_{0}(t) = 
    \begin{cases}
      \mathbf{q}_i + (\mathbf{g} - \mathbf{q}_i)f_{MJT}(t) & \text{ $0\le t < T$} \\
      \mathbf{g} & \text{ $T \le t$}
    \end{cases}
    \\
    & \quad\quad\quad f_{MJT}(t) = 10\Big(\frac{t}{T}\Big)^{3} - 15\Big(\frac{t}{T}\Big)^{4} + 6\Big(\frac{t}{T}\Big)^{5}
\end{split}     
\end{align}

\begin{figure}
\centering
\includegraphics[trim={2.0cm 2.0cm 3.0cm 2.0cm}, width=1.00\columnwidth, clip, page=1]{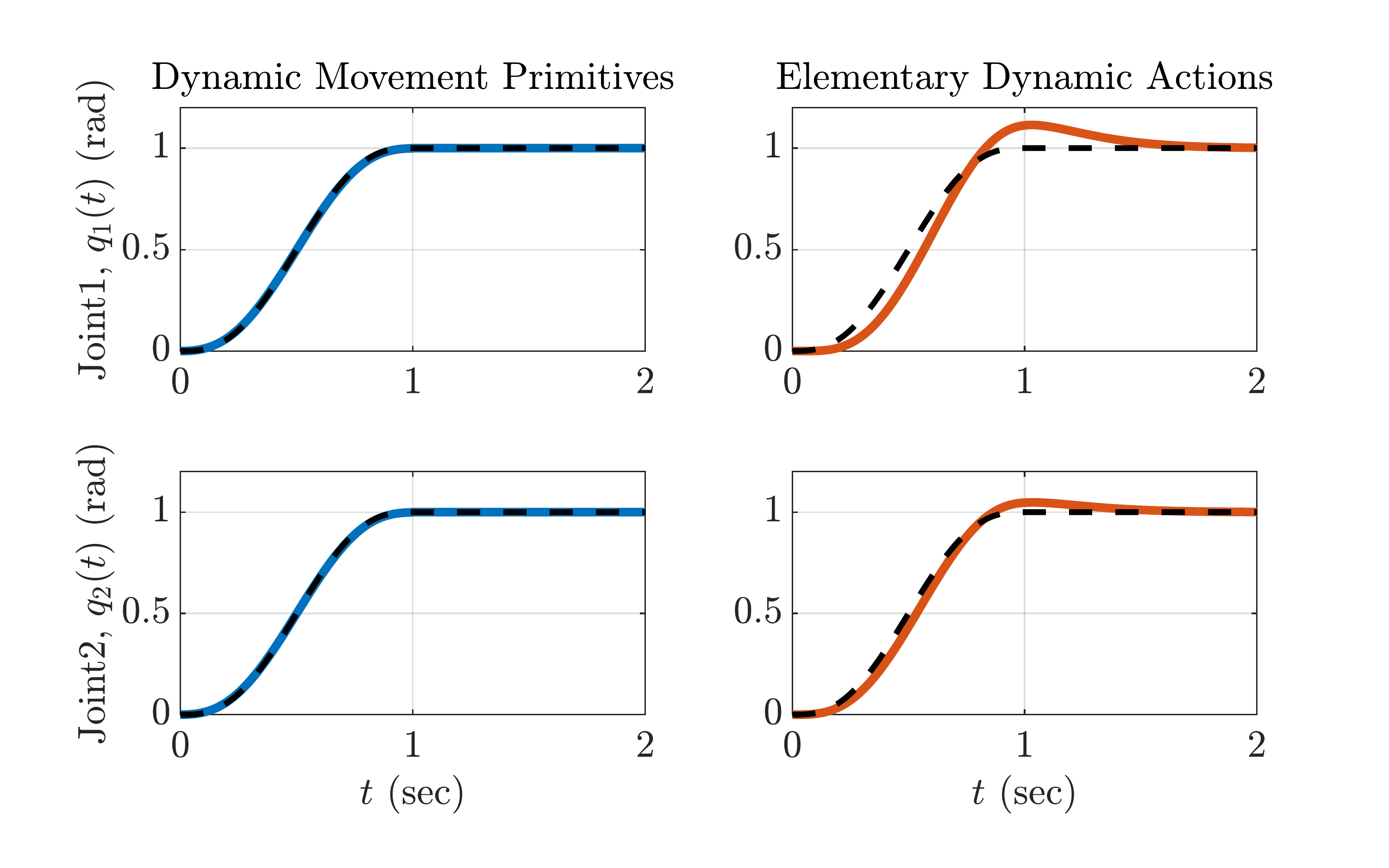}
\caption{ A goal-directed discrete movement in joint-space for Dynamic Movement Primitives (DMPs, blue) and Elementary Dynamic Actions (EDAs, orange) (Section \ref{subsubsec:joint_space_discrete_simulation_example}). The black dotted line is a minimum-jerk trajectory (Equation \eqref{eq:minimum_jerk_trajectory}). Goal location: $\mathbf{g}=[1, 1]$ rad. Parameters of minimum-jerk trajectory: $\mathbf{q}_i=[0,0]$ rad, $T=1$s. Parameters of DMPs: $\alpha_z=10$, $\beta_z=2.5$, $\alpha_s=1.0$, $\tau=T$, $N=50$, $P=100$, $c_i = \exp( -\alpha_s (i - 1)/ ( N-1) )$ for $i\in[1,2,\cdots, N]$, $h_i=1/(c_{i+1} - c_i)^2$ for $i\in[1,2,\cdots, N-1]$, $h_{N}=h_{N-1}$. Parameters of EDAs: $\mathbf{K}_q=150\mathbf{I}_2$ N$\cdot$m/rad, $\mathbf{B}_q=50\mathbf{I}_2$ N$\cdot$m$\cdot$s/rad, where $\mathbf{I}_2\in\mathbb{R}^{2\times 2}$ is an identity matrix. For DMP, perfect tracking was achieved. For EDA, a non-negligible tracking error was observed.
}
\label{fig:dmps_vs_edas_discrete_joint}        
\end{figure}

The simulation results are shown in Figure \ref{fig:dmps_vs_edas_discrete_joint}. 
DMPs generated the goal-directed discrete movement while achieving perfect tracking of the minimum-jerk trajectory.
Once the weights of the nonlinear forcing terms are learned using Imitation Learning, one can regenerate the minimum-jerk trajectory by simply retrieving these learned weights. 
While the presented example used a minimum-jerk trajectory, Imitation Learning can be used to achieve tracking control of a trajectory with arbitrary complexity.

For EDAs, a non-zero error between $\mathbf{q}_0(t)$ and $\mathbf{q}(t)$ was observed. 
Hence, perfect tracking of the minimum-jerk trajectory was not achieved.
Nevertheless, EDAs enabled asymptotic convergence of $\mathbf{q}(t)$ towards the goal configuration $\mathbf{g}$ without the need for an inverse dynamics model.

\subsection{A Goal-directed Discrete Movement in Task-space}\label{subsec:task_space_traj_track_no_redund}
We next design a controller to generate a goal-directed discrete movement of the end-effector in task-space coordinates.
For this Section, we assume no kinematic redundancy of the robot model, i.e., the Jacobian matrix $\mathbf{J}(\mathbf{q})$ is a square matrix.
A method to manage kinematic redundancy is considered in Section \ref{subsec:managing_kin_redund}.
One of the goal positions is located at a kinematic singularity, i.e., a fully stretched configuration.

\begin{figure*}
    \centering
    \includegraphics[trim={1.0cm 2.0cm 1.0cm 1.0cm}, width=0.80\textwidth, clip, page=1]{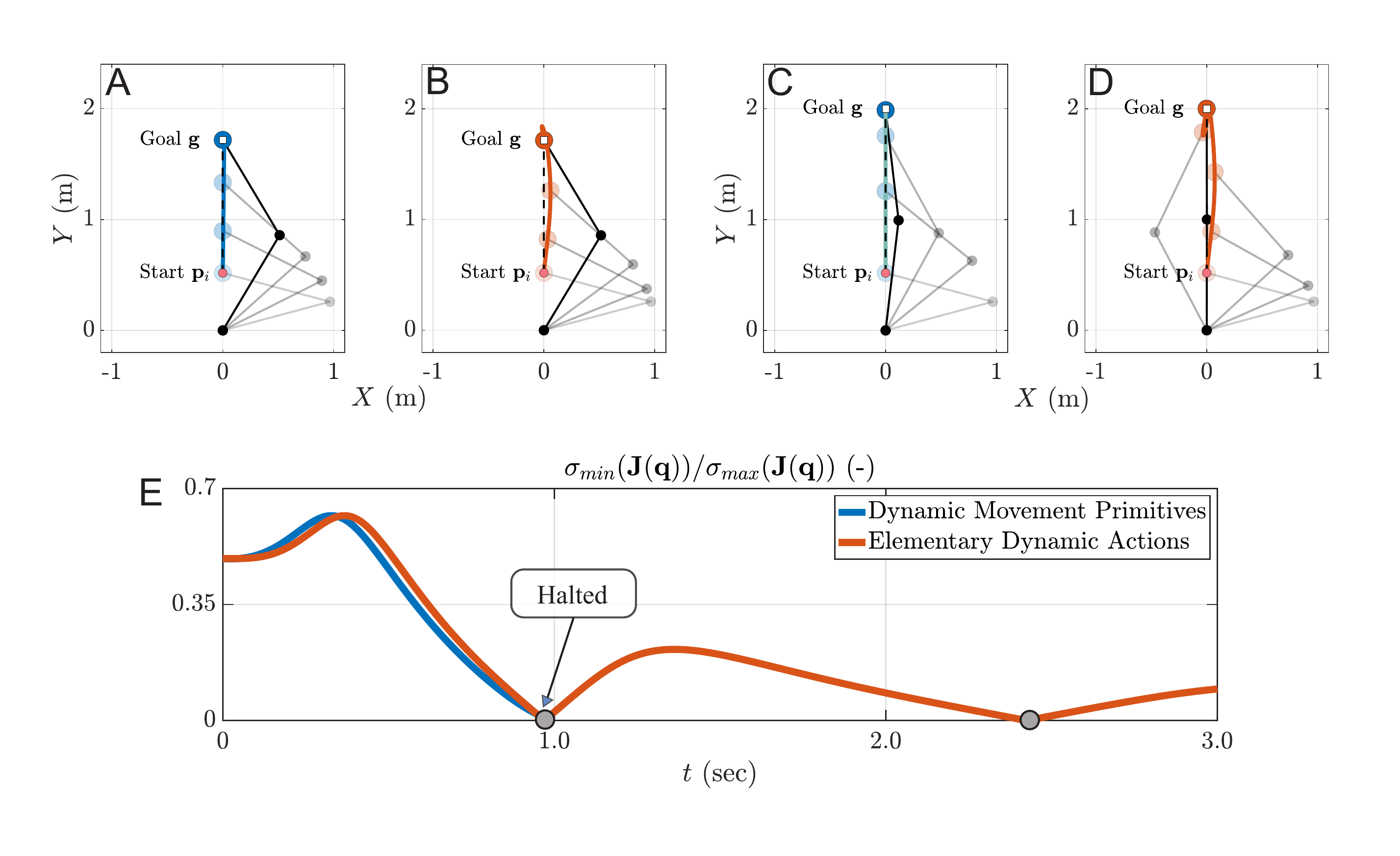}    
    \caption{ A goal-directed discrete movement in task-space for (A, C, E) Dynamic Movement Primitives (DMPs, blue) and (B, D, E) Elementary Dynamic Actions (EDAs, orange) (Section \ref{subsubsec:task_space_no_redund_discrete}). (A, B, C, D) Goal-directed discrete movements in a direction along the positive $Y$-axis direction, but (C, D) reached kinematic singularity. The black dotted line is a minimum-jerk trajectory (Equation \eqref{eq:minimum_jerk_trajectory_task}). 
    (E) Time $t$ vs. $\sigma_{min}((\mathbf{J(q)}))/\sigma_{max}((\mathbf{J(q)}))$ of both robots for (C, D), where $\sigma$ is a singular value of the matrix. For DMPs, the MuJoCo simulation halted near 0.99s.
    Parameters of a minimum-jerk trajectory: (A, B) $\mathbf{p}_{i}=[0.0, 0.52]$m, $\mathbf{g}=[0.0, 1.72]$m, $T=1.0$s (C, D) $\mathbf{p}_{i}=[0.0, 0.52]$m, $\mathbf{g}=[0.0, 2.00]$m, $T=1.0$s, where $\mathbf{p}_{i}$ is the initial end-effector position.
    Parameters of DMPs are identical to those in Figure \ref{fig:dmps_vs_edas_discrete_joint}.
    Parameters of EDAs: $\mathbf{K}_p=60\mathbf{I}_2$ N/m, $\mathbf{B}_p=20\mathbf{I}_2$ N$\cdot$s/m. For DMP, perfect tracking was achieved when the motion was not near a kinematic singularity. However for (C) when the goal $\mathbf{g}$ was at a kinematic singularty, the approach failed to achieve the task. For EDA, a non-negligible tracking error existed, but goal-directed discrete movements were achieved for both (A) and (C). The approach based on EDA was not only stable near and at a kinematic singularity but even passed through it (E, near $t=1.0$ and $t=2.4$, depicted as grey circle markers). }
    \label{fig:dmps_vs_edas_discrete_task}.   
\end{figure*}

\subsubsection{Dynamic Movement Primitives}\label{subsubsec:task_space_no_redund_dynamic_movement_primitives}
For DMPs, the task can be achieved by representing the end-effector trajectory $\mathbf{p}(t)$ with transformation system 
\citep{pastor2009learning}:
\begin{align}\label{eq:example_movement_primitives_nDOF_task_space}
    \begin{aligned}
    \tau^2\ddot{\mathbf{p}}(t) &= -\alpha_z\beta_z(\mathbf{p}(t)-\mathbf{g}) - \alpha_z\tau\dot{\mathbf{p}}(t) + \mathbf{f}(s(t))\\
          \tau\dot{s}(t) &= -\alpha_s s(t)
    \end{aligned}
\end{align}

Once the desired end-effector trajectories $\mathbf{p}_{des}$, $\dot{\mathbf{p}}_{des}$, $\ddot{\mathbf{p}}_{des}$ are computed with or without Imitation Learning (as discussed in Section \ref{subsec:joint_space_trajectory_tracking}), the inverse kinematics are used to calculate the corresponding joint trajectories:
\begin{align*}
    \mathbf{q}_{des} &= \mathbf{h}^{-1}(\mathbf{p}_{des})  \\
    \dot{\mathbf{q}}_{des} &= \mathbf{J}(\mathbf{q}_{des})^{-1}\dot{\mathbf{p}}_{des} \\
    \ddot{\mathbf{q}}_{des} &= \mathbf{J}(\mathbf{q}_{des})^{-1}\{ \ddot{\mathbf{p}}_{des} - \dot{\mathbf{J}}(\mathbf{q}_{des})\dot{\mathbf{q}}_{des}  \} 
\end{align*}
The calculated $\mathbf{\ddot{q}}_{des}, \mathbf{\dot{q}}_{des},\mathbf{q}_{des}$ are used as the input to the inverse dynamics model to calculate $\bm{\tau}_{in}$ (Section \ref{subsec:inverse_dynamics_model}).

Note that the presented method for inverse kinematics cannot be used for a kinematically redundant robot (with fewer task-space than joint-space DOFs).
To manage kinematic redundancy, along with the feedforward torque command from the inverse dynamics model, an additional feedback controller should be employed \citep{slotine1991applied, nakanishi2008operational, pastor2009learning}. This is considered further in (Section \ref{subsec:managing_kin_redund}). 

\subsubsection{Elementary Dynamic Actions}\label{subsubsec:discrete_task_motor_primitives}
To generate a goal-directed discrete movement planned in task-space coordinates, we construct the following controller:
\begin{equation}\label{eq:first_order_task_space_impedance}
    \bm{\tau}_{in}(t) = \mathbf{J}(\mathbf{q})^\text{T} \big[ \mathbf{K}_p \{\mathbf{p}_{0}(t) - \mathbf{p} \} + \mathbf{B}_p \{\dot{\mathbf{p}}_{0}(t) - \dot{\mathbf{p}} \} \big]
\end{equation}
In this equation, $\mathbf{K}_p, \mathbf{B}_p\in\mathbb{R}^{3\times 3}$ ($\mathbb{R}^{2\times 2}$ for the planar simulations presented below) are constant symmetric positive definite matrices which correspond to translational stiffness and damping, respectively.
This controller is a first-order task-space impedance controller \citep{hermus2021exploiting,verdi2019compositional}.

Compared to DMPs (Section \ref{subsubsec:task_space_no_redund_dynamic_movement_primitives}), this controller does not require a Jacobian inverse. Hence, the torque input is always well defined near and even at kinematic singularities.

To generate a goal directed discrete movement, as with the first-order joint-space impedance controller (Section \ref{subsec:joint_space_trajectory_tracking}), $\mathbf{p}_0(t)$ is chosen to be a submovement which ends at goal location $\mathbf{g}$.

For constant positive definite $\mathbf{K}_p, \mathbf{B}_p$ matrices, $\mathbf{p}(t)$ asymptotically converges to $\mathbf{g}$ \citep{takegaki1981new}.
This also implies an asymptotic convergence of $\mathbf{q}(t)\rightarrow\mathbf{h}^{-1}(\mathbf{g})$.

\subsubsection{Simulation Example}\label{subsubsec:task_space_no_redund_discrete}
The simulation example in Figure \ref{fig:dmps_vs_edas_discrete_task} reproduced the movement of the experiment conducted in \cite{burdet2001central}.
The goal-directed discrete movement was made in a direction away from the robot base, along the positive $Y$-axis direction (Figure \ref{fig:dmps_vs_edas_discrete_task}).
The amplitude of the movement was varied such that one of the movements reached a fully stretched configuration, i.e., a configuration at kinematic singularity.
The code script for this simulation is \texttt{main\_task\_discrete.py}.

We used the 2-DOF planar robot model from Section \ref{subsec:joint_space_trajectory_tracking} that was constrained to move within the $XY-$plane.
For DMPs and EDAs, both $\mathbf{p}_{des}(t)$ and $\mathbf{p}_0(t)$ were chosen to be a minimum-jerk trajectory (Equation \eqref{eq:minimum_jerk_trajectory}):
\begin{align}\label{eq:minimum_jerk_trajectory_task}
\begin{split}
    & \mathbf{p}_{des}(t), \mathbf{p}_{0}(t) = 
    \begin{cases}
      \mathbf{p}_i + (\mathbf{g} - \mathbf{p}_i)f_{MJT}(t) & \text{ $0\le t < T$} \\
      \mathbf{g} & \text{ $T \le t$}
    \end{cases}
    \\
    & \quad\quad\quad f_{MJT}(t) = 10\Big(\frac{t}{T}\Big)^{3} - 15\Big(\frac{t}{T}\Big)^{4} + 6\Big(\frac{t}{T}\Big)^{5}    
\end{split}    
\end{align}
In this equation, $\mathbf{p}_i$ is the initial position of the end-effector.

As shown in Figure \ref{fig:dmps_vs_edas_discrete_task}A, \ref{fig:dmps_vs_edas_discrete_task}B, both approaches successfully generated discrete movements that converged to the desired goal location. 
As discussed in Section \ref{subsubsec:joint_space_discrete_simulation_example}, DMPs achieved the goal-directed discrete movement with perfect tracking. 
For EDAs, goal-directed discrete movement was achieved but a tracking error still existed. 

Note that EDAs did not require the Jacobian inverse.
The benefit of this property was emphasized when the planar robot model reached for a fully stretched configuration. 
As shown in Figure \ref{fig:dmps_vs_edas_discrete_task}C, \ref{fig:dmps_vs_edas_discrete_task}D, \ref{fig:dmps_vs_edas_discrete_task}E, DMPs became numerically unstable when the robot model approached a kinematic singularity. 
For EDAs, not only was the approach  stable, but the approach even ``passed-through''  the kinematic singularity (Figure \ref{fig:dmps_vs_edas_discrete_task}E, near $t=1.0$ and $t=2.4$, depicted as grey circle markers) without any numerical instability of the controller. Note that the robot model oscillated back and forth between its ``left-hand'' and ``right-hand'' configurations, passing through the singularity multiple times. This occurred because at the singular configuration the controller included no effective damping, hence no means to dissipate the angular momentum of the robot links. This oscillation may be suppressed by adding a non-zero joint-space damping, an example of impedance superposition (Section \ref{subsubsec:mechanical_impedances}, \ref{subsec:obstacle_avoidance}, \ref{subsec:managing_kin_redund}).

To manage kinematic singularity for DMPs, methods such as ``damped least-squares inverse'' \citep{nakamura1986inverse,chiaverini1994review} can be employed.
In principle, these methods manually revise near-zero singular values to strictly positive values to avoid a singular $\mathbf{J(q)}$ matrix. 
However, this results in an approximate rather than exact value of the inverse kinematics near kinematic singularity. 
This error in the joint trajectory propagates to the feedforward torque command from the inverse dynamics model.
Hence, an additional feedback control method such as joint-space PD control (Section \ref{subsec:modulatingMechImpedances}) or sliding mode control (Section \ref{subsec:managing_kin_redund}) should be employed to ensure stability and reasonable tracking performance.

\begin{figure*}
    \centering
    \includegraphics[trim={0.0cm 0.0cm 0.0cm 0.0cm}, width=1.00\textwidth, clip, page=1]{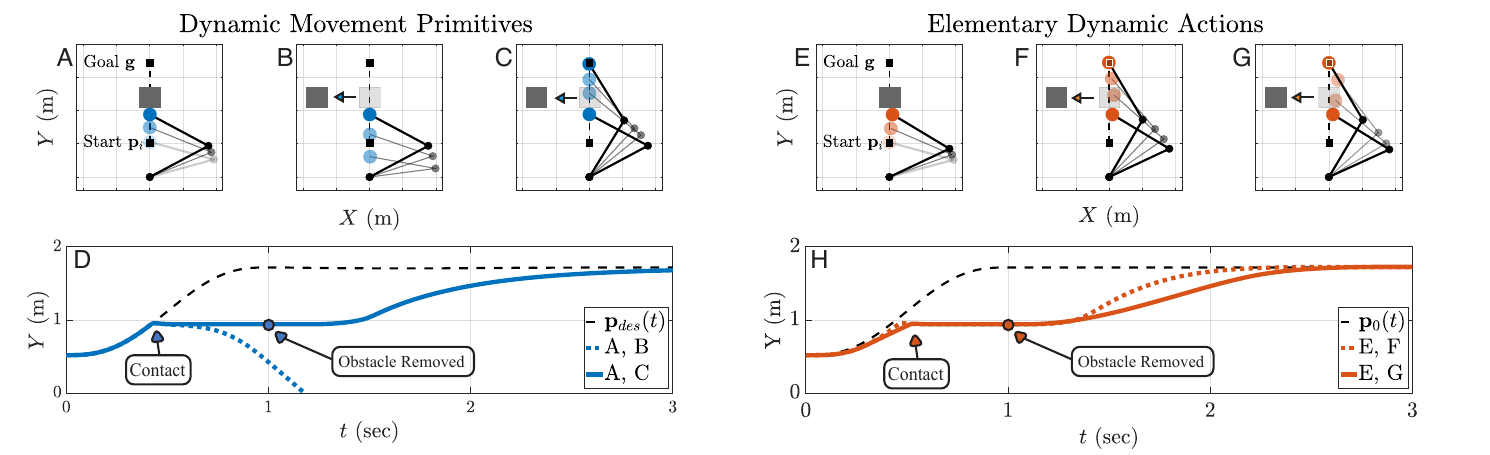}
    \caption{ A goal-directed discrete movement with unexpected physical contact for (A, B, C, D) Dynamic Movement Primitives (DMPs, blue) and (E, F, G, H) Elementary Dynamic Actions (EDAs, orange) (Section \ref{subsubsec:task_space_no_redund_discrete_unexpect}). For DMP: (A) Moment of contact. (A$\rightarrow$B) DMP without PD controller (A$\rightarrow$C) DMP with PD controller.
    (D) Time vs. $Y$-coordinate of the end-effector for A$\rightarrow$B (dotted line) and A$\rightarrow$C (filled line). For EDA: (E) Moment of contact. (E$\rightarrow$F) EDA without impedance modulation (E$\rightarrow$G) EDA with impedance modulation.
    (H) Time vs. $Y$-coordinate of the end-effector for E$\rightarrow$F (dotted line) and E$\rightarrow$G (filled line). With impedance modulation of E$\rightarrow$G, a slower movement than E$\rightarrow$F was achieved. (D, H) The black dotted lines are a minimum-jerk trajectory (Equation \eqref{eq:minimum_jerk_trajectory_task}).
    Parameters of the minimum-jerk trajectory, DMPs and EDAs are identical to those in Figure \ref{fig:dmps_vs_edas_discrete_task}. Parameters of the PD controller of DMPs (Equation \eqref{eq:dynamic_movement_primitives_superimposed}): $\mathbf{K}_q=50\mathbf{I}_2$ N$\cdot$m/rad, $\mathbf{B}_q=30\mathbf{I}_2$ N$\cdot$m$\cdot$s/rad. Smaller gain values than Figure \ref{fig:dmps_vs_edas_discrete_joint} were chosen to achieve a relatively low-gain PD controller as recommended by \citep{schaal2007dynamics,pastor2013dynamic}.
    Parameters of EDAs: $\mathcal{L}_{max}=2.5$J. }
    \label{fig:unexpected_contact}    
\end{figure*}

\subsection{Tasks with Unexpected Physical Contact}\label{subsec:modulatingMechImpedances}
We next consider a case where unexpected physical contact is made while conducting a point-to-point discrete reaching movement presented in Section \ref{subsubsec:task_space_no_redund_discrete} (Figure \ref{fig:dmps_vs_edas_discrete_task}A, \ref{fig:dmps_vs_edas_discrete_task}B).

\subsubsection{Dynamic Movement Primitives}\label{subsec:modulatingMechImpedances_movement}
As discussed in Section \ref{subsec:inverse_dynamics_model}, DMPs use a feedforward torque command calculated from the inverse dynamics model. 
To handle possible instabilities for control tasks involving unexpected contact, a feedback torque command is superimposed on the feedforward torque command.

Let $\bm{\tau}_{ff}(t)$ be the feedforward torque command (Section \ref{subsec:inverse_dynamics_model}, \ref{subsec:task_space_traj_track_no_redund}) to produce $\mathbf{p}_{des}(t)$.
A feedback torque command $\bm{\tau}_{fb}(t)$ based on a low-gain joint-space PD controller is superimposed on $\bm{\tau}_{ff}(t)$:
\begin{align}\label{eq:dynamic_movement_primitives_superimposed}
    \begin{split}
    \bm{\tau}_{fb}(t) &= \mathbf{K}_q \{\mathbf{q}_{des}(t) - \mathbf{q} \} + \mathbf{B}_q \{\dot{\mathbf{q}}_{des}(t) - \dot{\mathbf{q}} \} \\
    \bm{\tau}_{in}(t) &= \bm{\tau}_{ff}(t) + \bm{\tau}_{fb}(t) 
    \end{split}
\end{align}
The gain values for $\mathbf{K}_q$ and $\mathbf{B}_q$ are manually chosen to be sufficiently small to manage unexpected contacts \citep{schaal2007dynamics}. 
Moreover, $\mathbf{q}_{des}(t), \dot{\mathbf{q}}_{des}(t)$ are derived by inverse kinematics of $\mathbf{p}_{des}(t)$, $\dot{\mathbf{p}}_{des}(t)$ (Section \ref{subsubsec:task_space_no_redund_dynamic_movement_primitives}).
The gains $\mathbf{K}_q$ and $\mathbf{B}_q$ can also be determined by stochastic optimal control \citep{theodorou2010generalized,buchli2011learning}. 

Note that for ideal torque-actuated robots, the joint-space PD controller (Equation \eqref{eq:dynamic_movement_primitives_superimposed}) is identical to the first-order joint-space impedance controller of EDAs (Equation \eqref{eq:joint_space_impedance_controller}). 
However, it would be a mistake to conclude that PD control is identical to impedance control \citep{won1997comment}. Further discussion is deferred to Section \ref{subsubsec:contact_and_physical_interaction}.

\subsubsection{Elementary Dynamic Actions}\label{subsubsec:modulatingMechImpedances_motor}
As discussed in \cite{hogan1985impedance} and \cite{hogan2022contact}, an impedance controller is robust against unexpected physical contact with passive environments. 
While it is common to use constant mechanical impedances, mechanical impedance can be modulated to regulate the dynamics of physical interaction \citep{lachner_energy_2021}. 

In detail, the first-order task-space impedance controller of Equation \eqref{eq:first_order_task_space_impedance} can be adapted by modulating the translational stiffness and damping values:
\begin{equation}\label{eq:task_space_impedance_modulated}
    \bm{\tau}_{in} = \mathbf{J}(\mathbf{q})^\text{T} \big[ \mathbf{K}_p'(\lambda) \{\mathbf{p}_{0} - \mathbf{p} \} + \mathbf{B}_p'(\lambda) \{\dot{\mathbf{p}}_{0} - \dot{\mathbf{p}} \} \big]
\end{equation}
In this equation, $\mathbf{K}_p'(\lambda)=\lambda \mathbf{K}_p$ and $\mathbf{B}_p'(\lambda)=c\lambda \mathbf{K}_p$; $\mathbf{K}_p$ is a constant symmetric positive definite matrix; $\lambda \in [0,1]$, and $c$ is a positive constant that determines the damping ratio.

With a slight abuse of notation, $\mathbf{K}_p'(\lambda)$ and $\mathbf{B}_p'(\lambda)$ were modulated via $\lambda \equiv \lambda(\mathcal{T},\mathcal{U},\mathcal{L}_{max})$, which is a function of the kinetic energy of the robot $\mathcal{T}\equiv\mathcal{T}(\mathbf{q}(t), \dot{\mathbf{q}}(t))$, elastic potential energy $\mathcal{U}\equiv\mathcal{U}(\Delta \mathbf{p}(t), \mathbf{K}_p )$ due to the translational stiffness $\mathbf{K}_p$ and displacement $\Delta \mathbf{p}(t)$, and the energy threshold $\mathcal{L}_{max}$:
\begin{equation*}
\lambda=
   \begin{dcases}
     \hfil 1 & \text{if} \ \mathcal{L}_c(t,\lambda) \leq \mathcal{L}_{\textrm{max}} \\
     \hfil \max \bigg( \frac{1}{\mathcal{U}}(\mathcal{L}_{\textrm{max}} - \mathcal{T}), ~~ 0 \bigg) & \text{if} \ \mathcal{L}_c(t,\lambda) > \mathcal{L}_{\textrm{max}} 
   \end{dcases}
\end{equation*}
where:
\begin{align*}
   \mathcal{T}(\mathbf{q}, \dot{\mathbf{q}}) &= \frac{1}{2}\dot{\mathbf{q}}^{\text{T}}\mathbf{M}(\mathbf{q})\dot{\mathbf{q}} \\ 
   \mathcal{U}(\Delta \mathbf{p}, \mathbf{K}_p ) &= \frac{1}{2}\Delta \mathbf{p}^{\text{T}} \mathbf{K}_p \Delta \mathbf{p} \\
   \mathcal{L}_c(t,\lambda) &= \mathcal{T}(\mathbf{q}, \dot{\mathbf{q}}) + \lambda \mathcal{U}(\Delta \mathbf{p} ) 
\end{align*}

This controller limits the impact of an unexpected contact, e.g., during physical Human-Robot Interaction (pHRI) \citep{lachner2022geometric}, by bounding the total energy of the robot via $\lambda$.
For $\mathcal{T}\le\mathcal{L}_{\text{max}}$,
the total energy of the robot $\mathcal{L}_c$ is always less than or equal to $\mathcal{L}_{\text{max}}$. 
To avoid negative stiffness and damping matrices emerging from the condition $\mathcal{T}>\mathcal{L}_{\text{max}}$, $\lambda$ is set to be zero via the $\max$-function. 

For pHRI, the value $\mathcal{L}_{\text{max}}$ is defined by standards and regulations \citep{iso-ts_2016} and is dependent on the environment between the robot and the human. 

While the presented controller is a simplified example, a more advanced application exists where the damping term can be modulated as a function of stiffness and inertia \citep{albu-schaffer_cartesian_2003}.
Moreover, the dissipative behavior of the robot can be modulated to limit the robot power, e.g., by using ``damping injection'' \citep{stramigioli_energy-aware_2015}. 

\subsubsection{Simulation Example}\label{subsubsec:task_space_no_redund_discrete_unexpect}
As in Section \ref{subsubsec:task_space_no_redund_discrete}, we used a 2-DOF planar robot model to generate a goal-directed discrete movement in task-space coordinates. 
In this example, a square-shaped obstacle was placed to block the robot path.
A few seconds after the first contact with the robot, the obstacle was moved aside and the robot could continue its motion \citep{andrews1983impedance,newman1987high}. 
The code script for this simulation is \texttt{main\_unexpected\_contact.py}.

For DMPs, without a low-gain PD feedback controller, the learned weights from Section \ref{subsubsec:task_space_no_redund_discrete} were reused.
For this controller, the robot model bounced back from the obstacle due to contact and failed to reach the goal (Figure \ref{fig:unexpected_contact}B). 
Hence, it was necessary to add a low-gain PD controller (Figure \ref{fig:unexpected_contact}C).
The presented example showed the modular property of DMPs, since the learned feedforward torque controller from Section \ref{subsec:task_space_traj_track_no_redund} was reused without modification, and superimposed with an additional feedback controller.

\begin{figure}[H]
    \centering
    \includegraphics[trim={0.0cm 0.0cm 0.0cm 0.0cm}, width=0.99\columnwidth, clip, page=1]{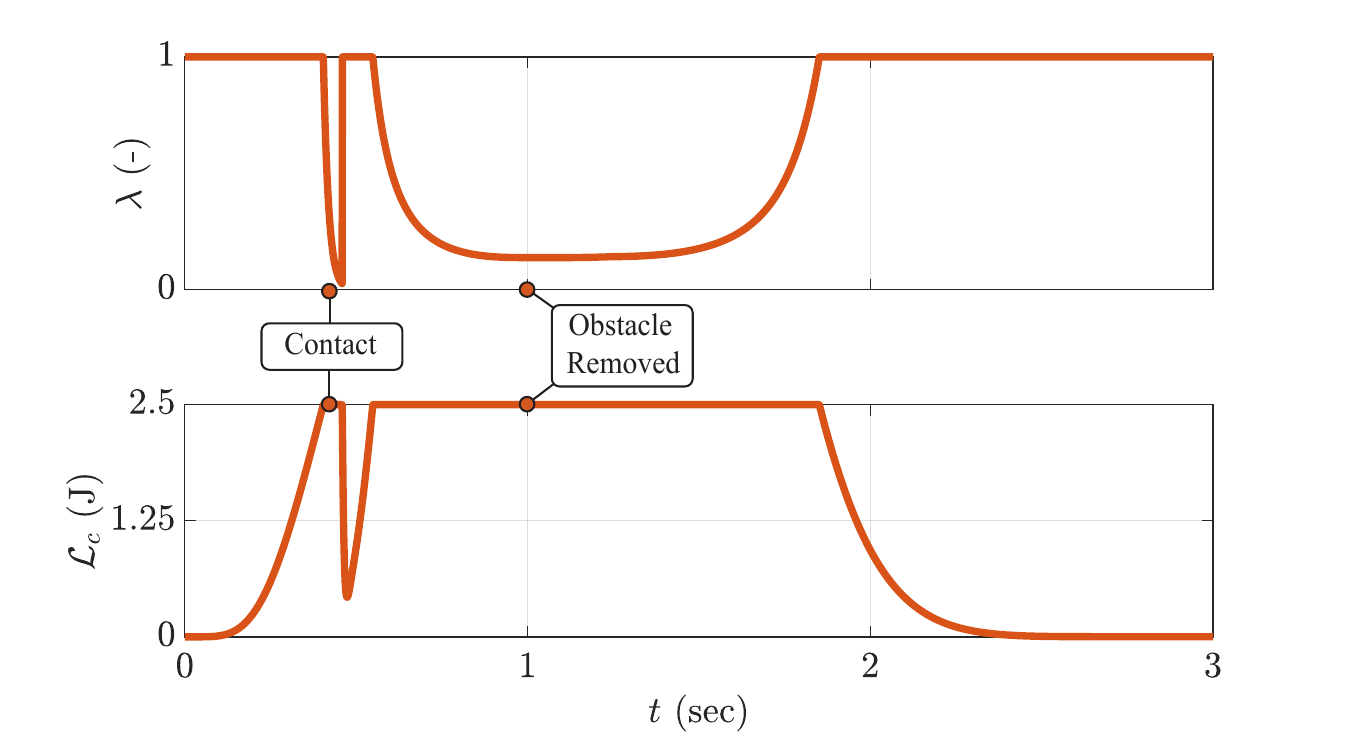}
    \caption{Simulation results using Elementary Dynamic Actions (EDAs) with impedance modulation via energy regulation (Equation \eqref{eq:task_space_impedance_modulated}) (Figure \ref{fig:unexpected_contact}) (Section \ref{subsubsec:task_space_no_redund_discrete_unexpect}). (Top) Time $t$ vs. $\lambda$ of the controller. After the first contact, a sudden change in $\lambda$ occurred. 
    (Bottom) Time vs. $\mathcal{L}_c$ of the controller. Since the zero-force trajectory continued during contact, the decrease of $\lambda$ limited $\mathcal{L}_c$ to a maximal value of $2.5$J. 
    As expected, $\mathcal{L}_c$ did not exceed $\mathcal{L}_{max}=2.5$J. 
    }
    \label{fig:unexpected_contact_motor_lambda}  
\end{figure}

For EDAs, both the controller with and without energy limitation were able to reach the goal (Figure \ref{fig:unexpected_contact}F, \ref{fig:unexpected_contact}G). 
However in the latter case, high accelerations of the end-effector occured after the obstacle was removed (Figure \ref{fig:unexpected_contact}H). 
As shown in Figure \ref{fig:unexpected_contact_motor_lambda}, $\lambda$ is regulated to limit the total energy to be less than $\mathcal{L}_{max}$.
By including mechanical impedance as a separate primitive, the EDAs approach was able to regulate the interactive dynamics between the robot and environment.

\subsection{Obstacle Avoidance}\label{subsec:obstacle_avoidance}
We next consider obstacle avoidance while conducting a point-to-point discrete reaching movement presented in Section \ref{subsubsec:task_space_no_redund_discrete}.
Moreover, we assume that the obstacle is fixed at location $\mathbf{o}\in\mathbb{R}^3$ ($\mathbb{R}^2$ for the planar simulations), although the method can be easily generalized to non-stationary obstacles. 

\subsubsection{Dynamic Movement Primitives}
To avoid an obstacle located at $\mathbf{o}\in\mathbb{R}^3$ 
a coupling term $\mathbf{C}_t(t)$ is added to the transformation system \citep{hoffmann2009biologically, ijspeert2013dynamical}:
\begin{align*}
    \tau^2\ddot{\mathbf{p}}(t) &= -\alpha_z\beta_z\{\mathbf{p}(t)-\mathbf{g}\} - \alpha_z\tau\dot{\mathbf{p}}(t) + \mathbf{f}(s(t)) + \mathbf{C}_t(t)
\end{align*}
The coupling term $\mathbf{C}_t(t)$ is defined by:
\begin{align}\label{eq:movement_primitives_coupling_term}
\begin{split}
    \mathbf{C}_t(t) &= \gamma \mathbf{R}(t)\dot{\mathbf{p}}(t)\theta(t) \exp(-\beta \theta(t)) \\
    \theta(t) &= \arccos{\bigg( \frac{ \{ \mathbf{o} - \mathbf{p}(t)\}^{\text{T}}\dot{\mathbf{p}}(t) }{|| \mathbf{o} - \mathbf{p}(t) || ~ ||\dot{\mathbf{p}}(t) || } \bigg) } 
\end{split}    
\end{align}
In these equations, $\gamma$ and $\beta$ are positive constants which correspond to the amplitude of the coupling term and its exponential decay rate, respectively;
$\theta(t)$ is the angle between $\mathbf{\dot{p}}(t)$ and $\mathbf{o-p}(t)$; $||\cdot ||:\mathbb{R}^3\rightarrow \mathbb{R}_{\ge0}$ 
is the Euclidean norm operator; $\mathbf{R}(t)\in\mathbb{R}^{3\times3}$ 
is a rotation matrix representing the $\pm90^\circ$ rotation about axis $\mathbf{r}(t)\equiv \{\mathbf{o}-\mathbf{p}(t) \}\times\dot{\mathbf{p}}(t)$.

For spatial task, $\mathbf{R}(t)$ can be calculated using the Rodrigues' formula \citep{murray1994mathematical}:
\begin{align*}
    \mathbf{R}(t) &= \mathbf{I}_3 + \sin\big( \pm \frac{\pi}{2} \big)  [\hat{\mathbf{r}}(t)] + \Big\{ 1 - \cos\big( \pm \frac{\pi}{2} \big) \Big\} [\hat{\mathbf{r}}(t)]^2 \\
    &= \mathbf{I}_3 \pm [\hat{\mathbf{r}}(t)] + [\hat{\mathbf{r}}(t)]^2
\end{align*}
In this equation, $\hat{\mathbf{r}}(t)$ is the normalization of $\mathbf{r}(t)$; $[\hat{\mathbf{r}}(t)]$ is a skew-symmetric matrix form of $\hat{\mathbf{r}}(t)$ \citep{murray1994mathematical}. The $+$ and $-$ signs represent $+90^{\circ}$ and $-90^{\circ}$ rotation about axis $\hat{\mathbf{r}}(t)$, respectively.
Intuitively, the coupling term forces the movement to move away from the obstacle \citep{ijspeert2013dynamical}. 
For a planar task, $\mathbf{R}(t)$ is a skew-symmetric matrix with $\pm 1$ and $\mp 1$ as off-diagonal terms.

\subsubsection{Elementary Dynamic Actions}
For EDAs, the idea resembles the method of obstacle avoidance using potential fields \citep{andrews1983impedance,newman1987high,hogan1985impedance,khatib1985real, koditschek1987exact,hjorth2020energy}.
A mechanical impedance which produces a repulsive force from the obstacle (i.e., a mechanical impedance with a point ``repeller'' at the obstacle location) is superimposed on the task-space impedance controller:
\begin{align}\label{eq:repulsive_impedance_ctrl}
    \mathbf{Z}_{p, 1}(t) &= \mathbf{K}_p \{\mathbf{p}_0(t) - \mathbf{p}(t) ) + \mathbf{B}_p \{ \dot{\mathbf{p}}_0(t) - \dot{\mathbf{p}}(t) \} \nonumber \\
    \mathbf{Z}_{p, 2}(t) &= - \frac{k}{|| \mathbf{o} - \mathbf{p}(t) ||^n} ( \mathbf{o} - \mathbf{p}(t) ) \nonumber \\
    \bm{\tau}_{in}(t) &= \mathbf{J}(\mathbf{q})^\text{T} \{ \mathbf{Z}_{p, 1}(t) + \mathbf{Z}_{p, 2}(t) \}
\end{align}
In these equations, $k$ is a positive constant that determines the amplitude of the repulsive force; $n$ is a positive integer. 
Intuitively, the impedance term $\mathbf{Z}_{p, 2}(t)$ produces a ``potential barrier'' which repels the robot from the obstacle.

\subsubsection{Simulation Example}\label{subsubsec:obstacle_avoid_simulation}
As in Section \ref{subsubsec:task_space_no_redund_discrete}, we used the 2-DOF planar robot model to generate a goal-directed discrete movement in task-space coordinates. 
However, a stationary obstacle was located at $\mathbf{o}$, blocking the path. 
The code script for this simulation is \texttt{main\_obstacle\_avoidance.py}.

\begin{figure}
    \centering
    \includegraphics[trim={0.0cm 1.0cm 2.0cm 2.0cm}, width=1.0\columnwidth, clip, page=1]{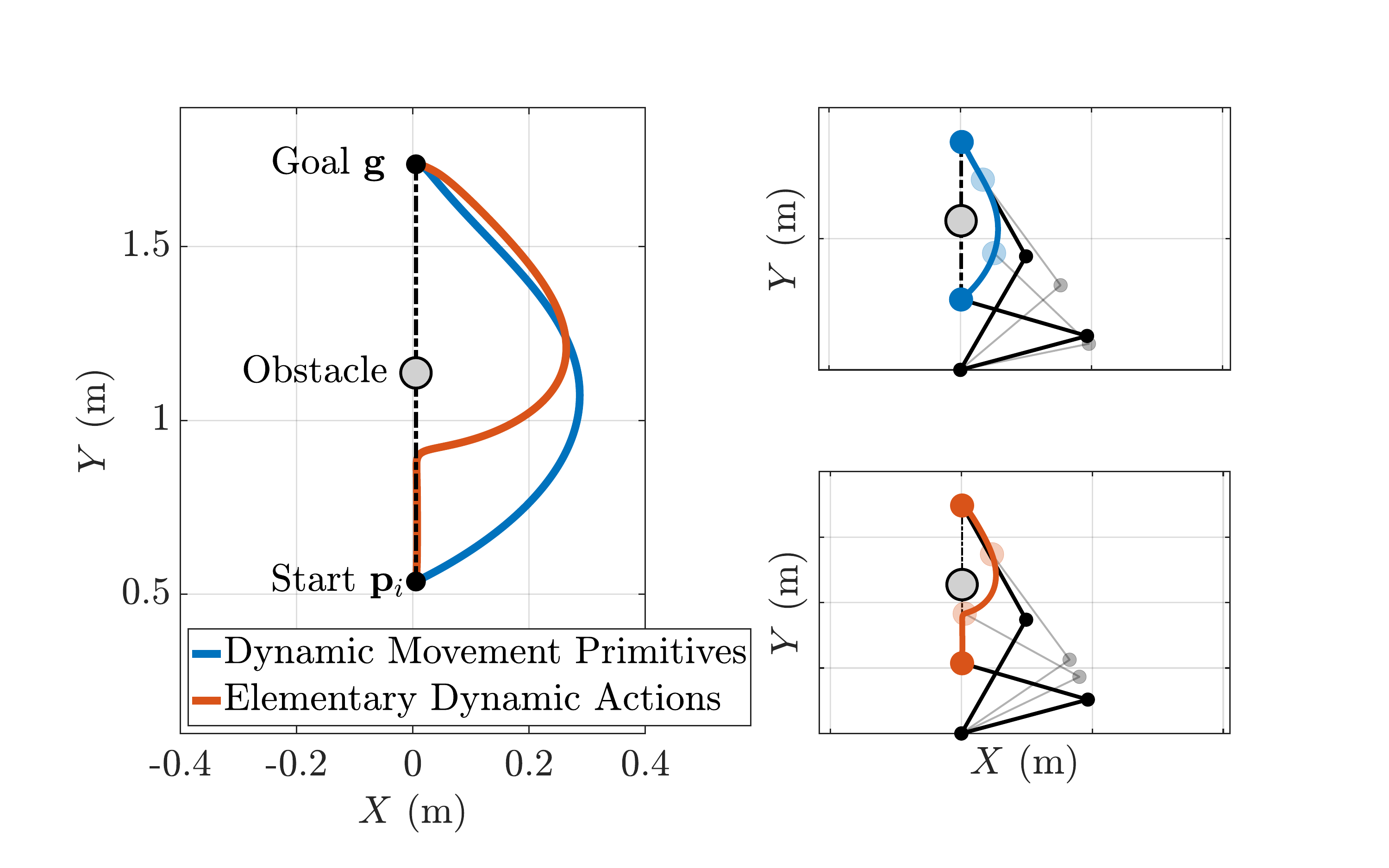}
    \caption{ A goal-directed discrete movement with obstacle avoidance for Dynamic Movement Primitives (DMPs, blue) and Elementary Dynamic Actions (EDAs, orange) (Section \ref{subsubsec:obstacle_avoid_simulation}). (Left) Trajectory of the end-effector position. Initial end-effector position $\mathbf{p}_i$ and goal location $\mathbf{g}$ are depicted as black circle markers. (Right) Time-lapse of the movement for DMP (Top, blue) and EDA (Bottom, orange). Dotted lines are a minimum-jerk trajectory. Obstacle location $\mathbf{o}=[0.0, 1.14]$m. Parameters of DMPs: $\gamma=300$, $\beta=-3$ (Equation \eqref{eq:movement_primitives_coupling_term}). Parameters of EDAs: $k=0.1$N/m$^5$, $n=6$ (Equation \eqref{eq:repulsive_impedance_ctrl}). Other parameters are identical to those in Figure \ref{fig:dmps_vs_edas_discrete_task}.}
    \label{fig:obstacle}    
\end{figure}    

As shown in Figure \ref{fig:obstacle}, both approaches successfully achieved the task.
Nevertheless, differences between the two approaches were observed.

DMPs considered the problem from a position control perspective. 
The coupling term $\mathbf{C}_t(t)$ (Equation \eqref{eq:movement_primitives_coupling_term}) controlled the acceleration of the end-effector position $\mathbf{p}(t)$ for obstacle avoidance.
As with Section \ref{subsec:modulatingMechImpedances}, the presented example also highlighted the modular property of DMPs. 
Without modification and reusing the learned weights from Section \ref{subsubsec:task_space_no_redund_discrete}, DMPs simply superimposed coupling term $\mathbf{C}_t(t)$ onto the transformation systems to modify the learned trajectory. 
This example shows the property of online trajectory modulation of DMPs, which is an advantage over spline methods (Section \ref{subsubsec:imitation_learning}).

In comparison, EDAs achieved obstacle avoidance without explicit path planning \citep{hogan1985impedance}.
Instead, both goal-reaching and obstacle avoidance tasks were achieved by using the superposition principle of mechanical impedances (Equation \eqref{eq:superposition_of_mechanical_impedances}). 
The modular property of the superposition principle enabled EDAs to divide the task into multiple sub-tasks, allocate an appropriate mechanical impedance for each sub-task, and then combine the mechanical impedances to solve the original task.
For this case, mechanical impedance $\mathbf{Z}_{p, 1}$ was used for the goal-directed discrete movement planned in task-space coordinates and mechanical impedance $\mathbf{Z}_{p, 2}$ was used for obstacle avoidance. 
As with the weights of DMPs, $\mathbf{Z}_{p, 1}$ was simply reused from Section \ref{subsubsec:task_space_no_redund_discrete} without any modification.

While the modular property of EDAs simplified the approach, care is required for implementation. 
Since EDAs do not explicitly plan the path, for Figure \ref{fig:obstacle}, a slight offset (2cm) in the positive $X$ direction was added to ``push away'' the robot to the counterclockwise direction \citep{andrews1983impedance}. Without this offset, the robot's end-effector could get stuck, or move in a clockwise direction which thereby resulted in collision between the robot links and the obstacle. Several methods, such as using time-varying potential fields, have been demonstrated to overcome these limitations. 
For brevity, further details are not reviewed here and the reader is referred to \cite{andrews1983impedance,newman1987high,khatib1986potential,hjorth2020energy}.

\subsection{Rhythmic Movement}\label{subsec:rhythmic_movement}
We next consider a method to generate a rhythmic, repetitive movement.
For this example, we considered movements planned in joint-space and task-space respectively.

\subsubsection{Dynamic Movement Primitives}
For DMPs, we used a canonical system and nonlinear forcing terms of a rhythmic movement (Section \ref{subsec:dynamic_movement_primitives}). From the generated $\mathbf{q}(t)$ (respectively $\mathbf{p}(t)$), the process discussed in Section \ref{subsec:joint_space_trajectory_tracking} (respectively Section \ref{subsec:task_space_traj_track_no_redund}) was conducted.

\subsubsection{Elementary Dynamic Actions}
For EDAs, we defined $\mathbf{q}_0(t)$ (respectively $\mathbf{p}_0(t)$) to be a rhythmic movement which we aimed to follow.
With this virtual trajectory, either a joint-space (Equation \eqref{eq:joint_space_impedance_controller}) or task-space impedance controller (Equation \eqref{eq:first_order_task_space_impedance}) was used.

\subsubsection{Simulation Example}\label{subsubsec:rhythmic_movement_example}
We used the 2-DOF planar robot model from Section \ref{subsec:joint_space_trajectory_tracking} and Section \ref{subsec:task_space_traj_track_no_redund}.
The code scripts for the simulations are \texttt{main\_joint\_rhythmic.py} for joint-space and \texttt{main\_task\_rhythmic.py} for task-space.

The rhythmic movement in joint-space followed a sinusoidal trajectory. For DMPs and EDAs, both $\mathbf{q}_{des}(t)$ and $\mathbf{q}_{0}(t)$ were defined by:
\begin{equation}\label{eq:figure_sinusoidal_traj}
    \mathbf{q}_{des}(t), \mathbf{q}_{0}(t) = \mathbf{q}_{i} + \mathbf{q}_{A} \sin( \omega_0 t)
\end{equation}
In this equation, $\mathbf{q}_{A}\in\mathbb{R}^{2}$ is the amplitude of the rhythmic movement.

The rhythmic movement in task-space followed a circular trajectory. For DMPs and EDAs, both $\mathbf{p}_{des}(t)$ and $\mathbf{p}_{0}(t)$ were defined by:
\begin{equation}\label{eq:figure_circular_trajectory}
    \mathbf{p}_{des}(t), \mathbf{p}_{0}(t) = \mathbf{c} + [ r_0\cos(\omega_0 t), r_0\sin( \omega_0 t ) ]
\end{equation}
In this equation, $\mathbf{c}\in\mathbb{R}^{2}$ is the center location of the circle; $r_0$ is the radius of the circle.

\begin{figure}
        \centering
         \includegraphics[trim={0.0cm, 0.0cm, 0.0cm, 0.0cm},
         width=0.99\columnwidth, clip]{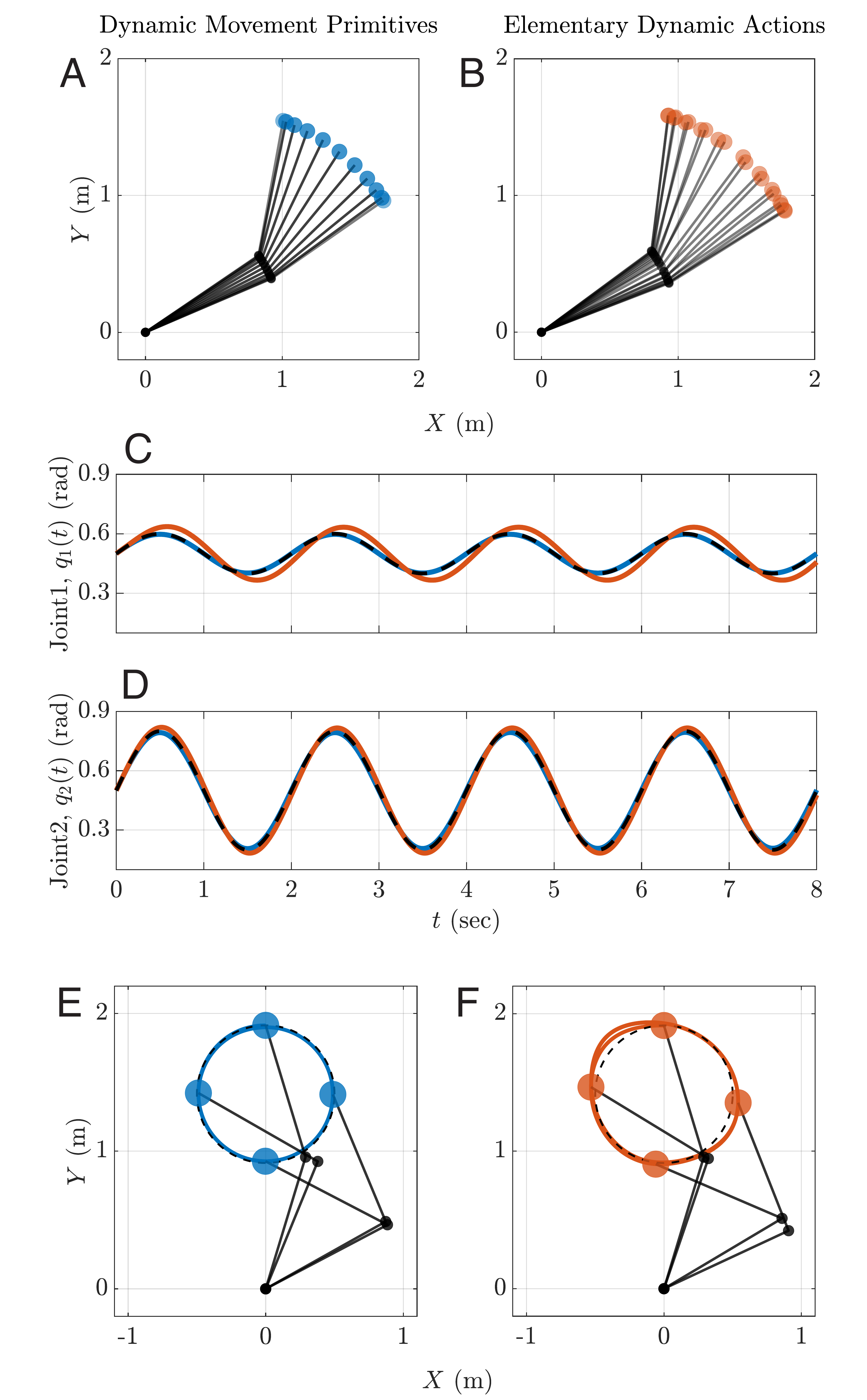} 
         \caption{ Rhythmic movements in (A, B, C, D) joint-space and (E, F) task-space for Dynamic Movement Primitives (DMPs, blue) and Elementary Dynamic Actions (EDAs, orange) (Section \ref{subsubsec:rhythmic_movement_example}). (A, B) Rhythmic movement in joint-space and its (C, D) joint trajectories. (C, D) The black dashed lines which are perfectly overlapped with DMPs (blue lines) represent a sinusoidal trajectory with parameters: $\mathbf{q}_i=[0.5, 0.5]$ rad, $\mathbf{q}_A=[0.1, 0.3]$ rad, $\omega_0=\pi$ rad/s (Equation \eqref{eq:figure_sinusoidal_traj}). Parameters of rhythmic DMP: $\alpha_z=10$, $\beta_z=2.5$, $\tau=1.0$, $r=1.0$, $N=40$, $P=100$, $c_i = 2\pi (i-1)/N$, $h_i=N$ for $i\in[1,2,\cdots,N]$. Parameters of EDA are identical to those in Figure \ref{fig:dmps_vs_edas_discrete_joint}. (E, F) Rhythmic movement in task-space. The black dashed lines represent a circular trajectory with parameters: $\mathbf{c}=[0, 1.4142]$m, $r_0=0.5$m, $\omega_0=\pi$ rad/s (Equation \eqref{eq:figure_circular_trajectory}).
         Parameters of EDAs:  $\mathbf{K}_p=90\mathbf{I}_2$ N/m, $\mathbf{B}_p=60\mathbf{I}_2$ N$\cdot$s/m. Higher stiffness and damping values than Figure \ref{fig:dmps_vs_edas_discrete_task} were chosen to reduce the tracking error.
         Parameters of rhythmic DMP are identical to those in (A, B). Consistent with Figure \ref{fig:dmps_vs_edas_discrete_joint} and Figure \ref{fig:dmps_vs_edas_discrete_task}, for DMP, perfect tracking was achieved while for EDA, a non-negligible tracking error existed.
 } \label{fig:rhythmic_joint_and_task}
\end{figure}

As shown in Figure \ref{fig:rhythmic_joint_and_task}, both DMPs and EDAs successfully generated rhythmic movement in joint-space (Figure \ref{fig:rhythmic_joint_and_task}A, \ref{fig:rhythmic_joint_and_task}B, \ref{fig:rhythmic_joint_and_task}C, \ref{fig:rhythmic_joint_and_task}D) and task-space (Figure \ref{fig:rhythmic_joint_and_task}E, \ref{fig:rhythmic_joint_and_task}F) . 
As discussed in Section \ref{subsubsec:joint_space_discrete_simulation_example} and Section \ref{subsubsec:task_space_no_redund_discrete}, for DMPs, perfect tracking was achieved in both joint-space and task-space. 
Given the period of the rhythmic movement (Section \ref{subsubsec:imitation_learning}), rhythmic trajectories with arbitrary complexity can be learned using Imitation Learning. For EDAs, tracking error existed in both joint-space and task-space. 
Nevertheless, EDAs generated a rhythmic, repetitive movement without an inverse dynamics model and without solving the inverse kinematics.

\subsection{Combination of Discrete and Rhythmic Movements}\label{subsec:superposition_of_discrete_rhythmic}
We next designed a controller to generate a combination of discrete and rhythmic movements \citep{hogan2007rhythmic}. 
For this example, we considered a movement planned in both joint-space and task-space.

\begin{figure*}
    \centering
     \includegraphics[trim={0.0cm 4.0cm 1.0cm 0.0cm},
     width=1.00\textwidth, clip]{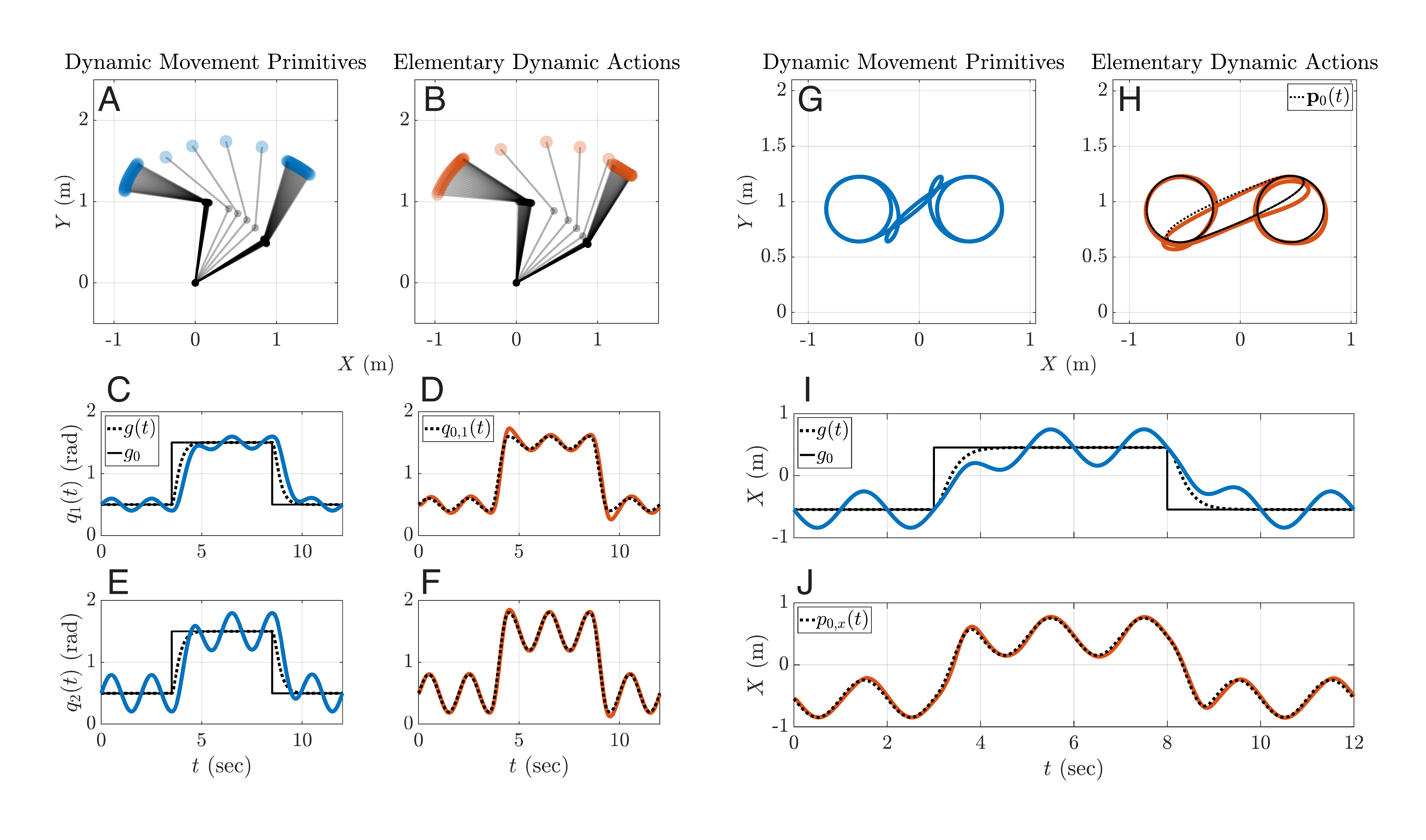} 
    \caption{A combination of discrete and rhythmic movements in (A-F) joint-space and (G-J) task-space for Dynamic Movement Primitives (DMPs, blue) and Elementary Dynamic Actions (EDAs, orange) (Section \ref{subsubsec:combination_of_discrete_rhythmic_example}). (C, E, I) Black filled lines represent the discrete change of $\mathbf{g}_0$ of DMP. Black dotted lines represent $\mathbf{g}(t)$, which follows a response of a (critically damped) second-order linear system. (D, F, J) Black dotted lines represent $\mathbf{q}_0(t)$ and $\mathbf{p}_0(t)$ for joint-space and task-space, respectively. (A-F) Parameters of the discrete and rhythmic movements in joint-space: $\omega_0=\pi$ rad/s, $\mathbf{q}_A=[0.1,0.3]$ rad, $\mathbf{q}_i=[0.5,0.5]$ rad, $\mathbf{q}_f=[1.5,1.5]$ rad. (G-J) Parameters of the discrete and rhythmic movements in task-space: $\omega_0=\pi$ rad/s, $r_0=0.3$m, $\mathbf{p}_i=[-0.47,0.9]$m, $\mathbf{p}_f=[0.53,0.9]$m. 
    In joint-space (respectively task-space), multiple discrete movements in opposite directions are generated by switching the values of $\mathbf{q}_i$ (respectively $\mathbf{p}_i$) and $\mathbf{q}_f$ (respectively $\mathbf{p}_f$), with time offset $t_{off}=3.5+5i$, where $i$ is the number of discrete movements.
    Parameters of discrete DMPs in both joint-space and task-space: $\tau_2=1.0$, $\alpha_{z,2}=10$, $\beta_{z,2}=2.5$. Other parameters of DMP are identical to those in Figure \ref{fig:rhythmic_joint_and_task}. Parameters of EDA are identical to those in Figure \ref{fig:rhythmic_joint_and_task}. For EDA, given a single impedance operator, discrete and rhythmic movements can be directly combined at the level of the virtual trajectory. }
    \label{fig:discrete_and_rhythmic}    
\end{figure*}

\subsubsection{Dynamic Movement Primitives}
For DMPs, the canonical system and nonlinear forcing term are different for discrete and rhythmic movements (Section \ref{subsubsec:canonical_system}, \ref{subsubsec:nonlinear_forcing_term}).
Hence, both rhythmic and discrete DMPs cannot be directly combined.
Instead, the discrete DMP generates a time-changing goal $\mathbf{g}(t)$ for the rhythmic DMP \citep{degallier2006movement,degallier2007hand,degallier2008modular}:
\begin{align*}
    \begin{aligned}
    \tau_1^2\ddot{\mathbf{q}}(t) &= -\alpha_{z,1}\beta_{z,1}\{\mathbf{q}(t)-\mathbf{g}(t)\} - \alpha_{z,1}\tau_1\dot{\mathbf{q}}(t) + \mathbf{f}(s(t))\\
          s(t) &=\frac{t}{\tau_1} ~~~~~~~~~~~~~~~ \mod 2\pi \\
    \tau_2^2\ddot{\mathbf{g}}(t) &= -\alpha_{z,2}\beta_{z,2}\{\mathbf{g}(t)-\mathbf{g}_0\} - \alpha_{z,2}\tau_2\dot{\mathbf{g}}(t)\\    
    \end{aligned}
\end{align*}
The first two equations represent rhythmic DMPs, and the last equation represents discrete DMPs without a nonlinear forcing term and a canonical system. $\mathbf{g}_0$ is the discontinuous change of the goal location to which goal $\mathbf{g}(t)$ converges. 
Note that for control in task-space, $\mathbf{q}(t)$ terms in the equation are substituted for $\mathbf{p}(t)$. 
From the generated $\mathbf{q}(t)$ (respectively $\mathbf{p}(t)$), the process discussed in Section \ref{subsec:joint_space_trajectory_tracking} (respectively Section \ref{subsec:task_space_traj_track_no_redund}) was conducted.

\subsubsection{Elementary Dynamic Actions}
For EDAs, we simply combine submovements $\mathbf{q}_{0, sub}(t)$ and oscillations $\mathbf{q}_{0, osc}(t)$ to define the virtual trajectory $\mathbf{q}_0(t)$:
\[
    \mathbf{q}_{0}(t) = \mathbf{q}_{0, sub}(t) + \mathbf{q}_{0, osc}(t) 
\]
Note that for control in task-space, $\mathbf{q}$ terms in the equation are substituted with $\mathbf{p}$.
With this virtual trajectory, either a joint-space (Equation \eqref{eq:joint_space_impedance_controller}) or a task-space impedance controller (Equation \eqref{eq:first_order_task_space_impedance}) is used.

\subsubsection{Simulation Example}\label{subsubsec:combination_of_discrete_rhythmic_example}
Using the 2-DOF robot model in Section \ref{subsec:joint_space_trajectory_tracking}, the goal was to generate a combination of discrete and rhythmic movements both in joint-space and task-space. 
The code scripts for the simulations are \texttt{main\_joint\_discrete\_and\_rhythmic.py} for joint-space and \texttt{main\_task\_discrete\_and\_rhythmic.py} for task-space.

For $\mathbf{q}_{des}(t)$ (respectively $\mathbf{p}_{des}(t)$) of DMPs, the sinusoidal trajectory, Equation \ref{eq:figure_sinusoidal_traj} (respectively circular trajectory, Equation \ref{eq:figure_circular_trajectory}) was represented by rhythmic DMPs, and $\mathbf{g}(t)$ of $\mathbf{q}_{des}(t)$ (respectively $\mathbf{p}_{des}(t)$) was represented by discrete DMPs with $\mathbf{g}_0$ discretely changing from $\mathbf{q}_i$ to $\mathbf{q}_{f}$ (respectively $\mathbf{p}_i$ to $\mathbf{p}_{f}$).

For $\mathbf{q}_{0}(t)$ of EDAs, a minimum-jerk trajectory (Equation \eqref{eq:minimum_jerk_trajectory}) was combined with a sinusoidal trajectory (Equation \eqref{eq:figure_sinusoidal_traj}):
\begin{align*}
    \mathbf{q}_{0}(t) &= \mathbf{q}_{0,sub}(t) + \mathbf{q}_{0,osc}(t) \\
    \mathbf{q}_{0,sub}(t) &=  \begin{cases}
        \mathbf{0} & \text{ $ 0 \le t < t_{off} $} \\
      \mathbf{q}_i + (\mathbf{q}_{f} - \mathbf{q}_i) f_{MJT}(t)& \text{ $ t_{off}\le t < T+t_{off}$} \\
      \mathbf{q}_{f} & \text{ $T+t_{off} \le t$}
    \end{cases} \\
    \mathbf{q}_{0,osc}(t) &= \mathbf{q}_A \sin(\omega_0 t) \\
    f_{MJT}(t) &= 10\Big(\frac{t-t_{off}}{T}\Big)^3 - 15\Big(\frac{t-t_{off}}{T}\Big)^4 + 6\Big(\frac{t-t_{off}}{T}\Big)^5   
\end{align*}
In this equation, $t_{off} >0$ is a time offset for the submovement.
For $\mathbf{q}_{des}(t)$ of DMPs, the sinusoidal trajectory was represented by rhythmic DMPs, and $\mathbf{g}(t)$ of $\mathbf{q}_{des}(t)$ was represented by discrete DMPs with $\mathbf{g}_0$ discretely changing from $\mathbf{q}_i$ to $\mathbf{q}_{f}$.

For $\mathbf{p}_{0}(t)$ of EDAs, a minimum-jerk trajectory (Equation \eqref{eq:minimum_jerk_trajectory}) was combined with a circular trajectory (Equation \eqref{eq:figure_circular_trajectory}):
\begin{align*}
    \mathbf{p}_{0}(t) &= \mathbf{p}_{0,sub}(t) + \mathbf{p}_{0,osc}(t) \\
    \mathbf{p}_{0,sub}(t) &=  \begin{cases}
        \mathbf{0} & \text{ $ 0 \le t < t_{off} $} \\
      \mathbf{p}_i + (\mathbf{p}_{f} - \mathbf{p}_i)f_{MJT, off}(t) & \text{ $ t_{off}\le t < T+t_{off}$} \\
      \mathbf{g} & \text{ $T+t_{off} \le t$}
    \end{cases} \\
    \mathbf{p}_{0,osc}(t) &= [ r_0\cos(\omega_0 t), r_0\sin( \omega_0 t ) ] 
\end{align*}

As shown in Figure \ref{fig:discrete_and_rhythmic}, both approaches successfully produced a combination of discrete and rhythmic movements in joint-space and task-space.
However, since DMPs separate the canonical system and nonlinear forcing terms for discrete and rhythmic movements (Section \ref{subsubsec:canonical_system}, \ref{subsubsec:nonlinear_forcing_term}), merging the two movements was not straightforward. 
DMPs circumvented this issue by assigning the time-changing goal $\mathbf{g}(t)$ of rhythmic DMPs, with discrete DMPs without the nonlinear forcing terms input.
Nevertheless, this resulted in $\mathbf{g}(t)$ that followed the response of a second-order linear system (Figure \ref{fig:discrete_and_rhythmic}). 

On the other hand for EDAs, given a single impedance operator, discrete and rhythmic movements were directly combined at the level of the virtual trajectory (i.e., the forward path dynamics) (Figure \ref{fig:edas_w_Norton_Network}).
With modest parameter tuning, the discrete and rhythmic movements used in Section \ref{subsubsec:joint_space_discrete_simulation_example} and Section \ref{subsubsec:rhythmic_movement_example} were reused and combined.
This approach intuitively provides convenience in practical implementation and also emphasizes the modularity of EDAs at a kinematic level. 
Moreover, the trajectory of the discrete movement need not be restricted to a response of a second-order linear system and can be freely chosen.

\subsection{Sequence of Discrete Movements}\label{subsec:goal_changing}
We next consider designing a controller to generate a sequence of discrete movements planned in task-space coordinates. 
For this, we show how the controller generates a movement in response to a sudden change of goal location.

\subsubsection{Dynamic Movement Primitives}
With the controller introduced in Section \ref{subsubsec:task_space_no_redund_dynamic_movement_primitives}, an additional differential equation for the time-varying goal location $\mathbf{g}(t)$ was added \citep{ijspeert2013dynamical}:
\begin{align}\label{eq:DMP_goal_changing_sequence}
    \begin{split}
    \tau^2\ddot{\mathbf{p}}(t) &= -\alpha_z\beta_z\{\mathbf{p}(t)-\mathbf{g}(t)\} - \alpha_z\tau\dot{\mathbf{p}}(t) + \mathbf{f}(s(t))\\
          \tau \dot{\mathbf{g}}(t) &= \alpha_g \{ \mathbf{g}_0 - \mathbf{g}(t) \}
    \end{split}
\end{align}
In these equations, $\alpha_g$ is a positive constant; $\mathbf{g}_0$ is a discontinuous change of the goal location. In other words, the goal location $\mathbf{g}$ is first-order low-pass filtered. This is used to avoid a discontinuous jump in the acceleration $\ddot{\mathbf{p}}(t)$ of the transformation system.

Note that both $\mathbf{p}(t)$ and $\mathbf{g}(t)$ comprise a third-order linear system with a nonlinear forcing term and $\alpha_g \mathbf{g}_0/\tau$ as inputs. 
While any positive values of $\alpha_g$ can be used, for $\tau=1$ and $\beta_z=\alpha_z/4$, $\alpha_g$ is often chosen to be $\alpha_g =\alpha_z/2$ such that the third-order linear system has three repeated eigenvalues \citep{nemec2012action,schaal2005learning}. 

Note that $\mathbf{g}(t)$ has a closed-form solution. 
Without loss of generality, if the goal location changes from $\mathbf{g}_0=\mathbf{g}_{old}$ to $\mathbf{g}_0=\mathbf{g}_{new}$ at time $t=0$, with initial condition $\mathbf{g}(0)=\mathbf{g}_{old}$:
\[
    \mathbf{g}(t) = \mathbf{g}_{new} + (\mathbf{g}_{old} - \mathbf{g}_{new} ) \exp\bigg({ -\frac{\alpha_g}{\tau} t} 
    \bigg)
\]
Hence, a sequence of finite submovements can be generated by discrete changes of the goal location from $\mathbf{g}_0=\mathbf{g}_{old}$ to $\mathbf{g}_0=\mathbf{g}_{new}$.  

While the first discrete movement can follow an arbitrary trajectory using Imitation Learning, the subsequent discrete movements cannot, but follow the motion of a stable third-order linear system converging to the corresponding $\mathbf{g}_{0}$ value.

\begin{figure*}
    \centering
    \includegraphics[trim={0.0cm 1.0cm 0.0cm 0.0cm},
     width=1.00\textwidth,clip]{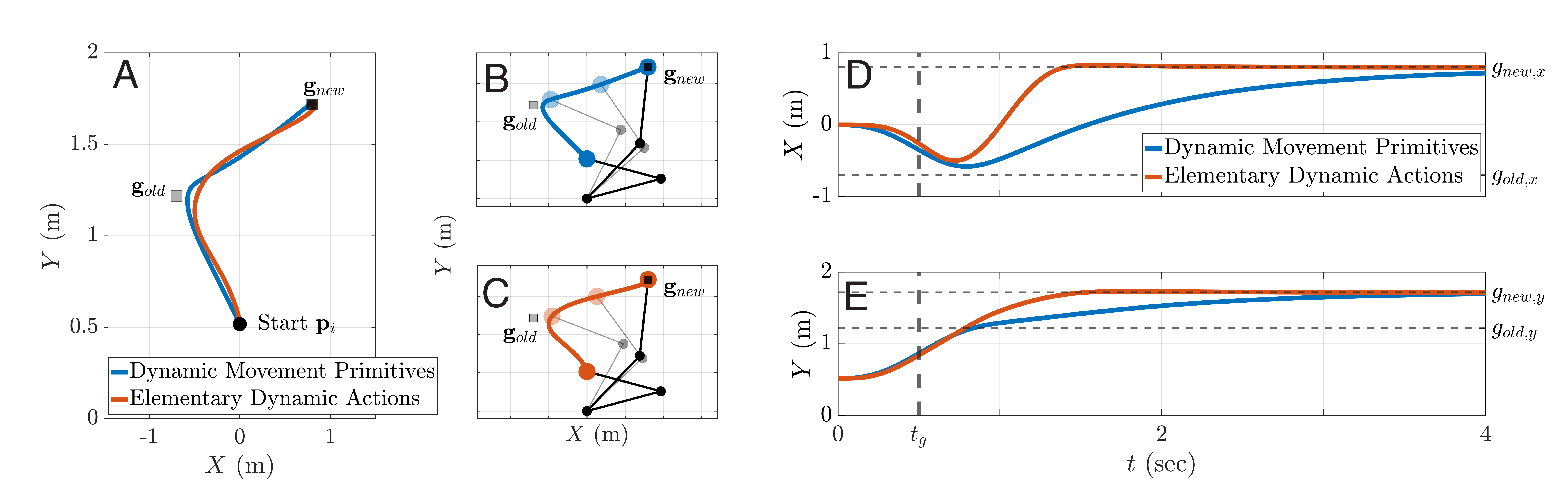}
    \caption{ A sequence of discrete movements for (A, B, D, E) Dynamic Movement Primitives (DMPs, blue) and (A, C, D, E) Elementary Dynamic Actions (EDAs, orange) (Section \ref{subsubsec:sequence_of_discrete_movements}). (A) Trajectory of the end-effector position. (B, C) Time-lapse of the movement of the 2-DOF robot model. (D, E) Time vs. $X$- and $Y$-coordinates of the end-effector trajectory. (A, B, C) The first movement headed toward $\mathbf{g}_{old}$ (square grey marker), but at time $t=t_g$ (D, E), the target switched location to $\mathbf{g}_{new}$ (square black marker) which necessitated a second movement. Parameters of DMPs (for the first discrete movement): $\alpha_g=1.0$, $\tau=1.0$. Other parameters are identical to those in Figure \ref{fig:dmps_vs_edas_discrete_task}. Parameters of EDAs: $T_1=1.0$s, $T_2=1.0$s, $t_g=0.5$s. $\mathbf{g}_{old}=[-0.7, 1.22]$m, $\mathbf{g}_{new}=[0.8, 1.72]$m, $\mathbf{p}_{i}=[0.0, 0.52]$m. Other parameters are identical to those in Figure \ref{fig:dmps_vs_edas_discrete_task}. (D, E) For the new goal location $\mathbf{g}_{new}$, EDA achieved faster convergence than DMP. For DMP the second movement followed a response of a third-order linear system. On the other hand for EDA, the basis function of the second movement can be freely chosen. Hence, faster convergence to $\mathbf{g}_{new}$ was achieved for EDA. }
    \label{fig:sequence}
\end{figure*}

\subsubsection{Elementary Dynamic Actions}
With the first-order task-space impedance controller (Equation \eqref{eq:first_order_task_space_impedance}), a sequence of discrete movements is generated by sequencing multiple submovements \citep{flash1991arm}:
\begin{align*}
    \dot{\mathbf{p}}_0(t) &= \sum_{i} \mathbf{v}_i \ \hat{\sigma}_i(t - t_i)
\end{align*}

The amplitude of the $i$-th basis function $\mathbf{v}_{i}$ is chosen to reach the goal of the $i$-th submovement, starting from the goal of the previous submovement. 
Note that each submovement can use a different basis function $\hat{\sigma}$. 

\subsubsection{Simulation Example}\label{subsubsec:sequence_of_discrete_movements}
The simulation example in Figure \ref{fig:sequence} reproduced the experiment conducted in \cite{flash1991arm}. Let the goal location of the first discrete movement be $\mathbf{g}_{old}$.
A new goal location, $\mathbf{g}_{new}$ suddenly appeared at time $t_g$.
The task was to modulate the controller to eventually reach $\mathbf{g}_{new}$.
We assumed that the new goal position $\mathbf{g}_{new}$ was immediately measured at time $t_g$.
The simulation example used the 2-DOF robot model of Section \ref{subsec:joint_space_trajectory_tracking}.
The code script for the simulation is \texttt{main\_sequencing.py} 

For DMPs, the first discrete movement followed a minimum-jerk trajectory with goal location $\mathbf{g}_{old}$.
The subsequent discrete movement was generated by discretely changing $\mathbf{g}_0= \mathbf{g}_{old}$ to $\mathbf{g}_0= \mathbf{g}_{new}$ at $t=t_g$:
\begin{align*}
    \begin{aligned}
        \mathbf{g}_0(t) &= 
        \begin{cases}
          \mathbf{g}_{old} & \text{ $0 \le t < t_g$ } \\
          \mathbf{g}_{new} & \text{ $t_g \le t$ } 
        \end{cases}         
    \end{aligned} 
\end{align*}

For EDAs, the minimum-jerk trajectory was used for the basis function of each submovement $\mathbf{p}_0(t)$:
\begin{align*}
        \mathbf{p}_0(t) &= \mathbf{p}_{0,1}(t) + \mathbf{p}_{0,2}(t)   \\
    \mathbf{p}_{0,1}(t) &= 
        \begin{cases}
          \mathbf{p}_i + (\mathbf{g}_{old} - \mathbf{p}_i)f_{MJT, 1}(t) & \text{ $0\le t < T_1$} \\
          \mathbf{g}_{old} & \text{ $T_1 \le t$}
        \end{cases}     \\
    \mathbf{p}_{0,2}(t) &= 
        \begin{cases}
            \mathbf{0} & \text{ $0\le t < t_g$}  \\  
          (\mathbf{g}_{new} - \mathbf{g}_{old})f_{MJT, 2}(t) & \text{ $t_g\le t < t_g + T_2$} \\
          \mathbf{g}_{new} - \mathbf{g}_{old} & \text{ $t_g + T_2 \le t$}
        \end{cases}  \\
f_{MJT, 1}(t) &= 10\Big(\frac{t}{T_1}\Big)^3 - 15\Big(\frac{t}{T_1}\Big)^4 + 6\Big(\frac{t}{T_1}\Big)^5  \\
f_{MJT, 2}(t) &= 10\Big(\frac{t-t_g}{T_2}\Big)^3 - 15\Big(\frac{t-t_g}{T_2}\Big)^4 + 6\Big(\frac{t-t_g}{T_2}\Big)^5 
\end{align*}
In these equations, subscripts $1$ and $2$ denote the first and second submovements, respectively.
With this $\mathbf{p}_0(t)$, the controller introduced in Section \ref{subsubsec:discrete_task_motor_primitives} was used.

As shown in Figure \ref{fig:sequence}, for both approaches, $\mathbf{p}(t)$ converged to the new goal location $\mathbf{g}_{new}$. 
For DMPs, the subsequent discrete movements by design followed the motion of a third-order linear system.
For EDAs, the subsequent submovements can use any basis functions, which thereby results in flexibility for determining the resulting motion.
As a result, EDAs were able to reach the $\mathbf{g}_{new}$ location faster than DMPs.
Moreover, the modular property of EDAs at the kinematics level enabled a smooth integration of the second movement without any modification of the first one.

\subsection{Managing Kinematic Redundancy}\label{subsec:managing_kin_redund}
We next consider designing a controller to generate a goal-directed (or sequence of) discrete movement(s) of the end-effector for a kinematically redundant robot.
By definition, kinematic redundancy occurs when a Jacobian matrix has a null space \citep{siciliano1990kinematic}.
Hence, infinitely many joint velocity solutions exist to produce a desired end-effector velocity.
While kinematic redundancy provides significant challenges, additional control objectives  can be achieved by exploiting kinematic redundancy. 
Examples include obstacle avoidance \citep{baillieul1986avoiding,maciejewski1985obstacle}, joint limit avoidance \citep{liegeois1977automatic, hjorth2020energy}, and minimization of instantaneous power during movement \citep{klein1983review}.

\subsubsection{Dynamic Movement Primitives}\label{subsubsec:task_space_redund_dynamic_movement_primitives}
For DMPs, a feedback controller is employed to manage kinematic redundancy.
Multiple feedback control methods exist and can be divided into three categories: velocity-based control, acceleration-based control, and force-based control \citep{nakanishi2005comparative,nakanishi2008operational}.
Within these methods, we used a ``velocity-based control without joint-velocity integration'' \citep{nakanishi2008operational,pastor2009learning}.

Let $\mathbf{p}_{des}(t)$ be the desired end-effector trajectory with duration $T$ and $\mathbf{p}_{des}(T)=\mathbf{g}$.
By generating $\mathbf{p}_{des}(t)$, $\dot{\mathbf{p}}_{des}(t)$, $\ddot{\mathbf{p}}_{des}(t)$ as shown in Section \ref{subsec:task_space_traj_track_no_redund}, a reference end-effector velocity $\dot{\mathbf{p}}_r(t)$ and its corresponding reference joint velocity $\dot{\mathbf{q}}_r(t)$ are defined:
\begin{align*}
    \dot{\mathbf{p}}_r(t) &= \dot{\mathbf{p}}_{des}(t) + \bm{\Lambda }_1(\mathbf{p}_{des}(t)- \mathbf{p}(t)) \\
    \dot{\mathbf{q}}_r(t)  &= \mathbf{J}(\mathbf{q})^{+}\dot{\mathbf{p}}_r(t) = \mathbf{J}(\mathbf{q})^{+}\big\{\dot{\mathbf{p}}_{des}(t) + \bm{\Lambda }_1(\mathbf{p}_{des}(t)- \mathbf{p}(t)) \big\}
\end{align*}
In these equations, $\mathbf{J}(\mathbf{q})^{+}$ denotes the Moore-Penrose pseudo inverse, which is defined by $\mathbf{J}(\mathbf{q})^{+}=\mathbf{J}(\mathbf{q})^{\text{T}}\{\mathbf{J}(\mathbf{q})\mathbf{J}(\mathbf{q})^{\text{T}}\}^{-1}$ \citep{penrose1955generalized,klein1983review};
$\bm{\Lambda}_1\in\mathbb{R}^{3\times 3}$ ($\mathbb{R}^{2\times 2}$ for planar task) is a symmetric positive definite matrix. 
Note that for $\dot{\mathbf{q}}_r(t)$, an additional term which projects an arbitrary joint velocity vector onto the null space of the Jacobian matrix $\mathbf{J}(\mathbf{q})$ can be added \citep{liegeois1977automatic}. 
However, this additional term was omitted for brevity. 

Accordingly, the reference end-effector acceleration $\ddot{\mathbf{p}}_r(t)$ and its corresponding reference joint acceleration $\ddot{\mathbf{q}}_r(t)$ are defined:
\begin{align*}
    \ddot{\mathbf{p}}_r(t) &= \ddot{\mathbf{p}}_{des}(t) + \bm{\Lambda }_1(\dot{\mathbf{p}}_{des}(t)- \dot{\mathbf{p}}(t)) \\
    \ddot{\mathbf{q}}_r(t)  &= \mathbf{J}(\mathbf{q})^{+}(\ddot{\mathbf{p}}_r(t) -\mathbf{\dot{J}}(\mathbf{q})\dot{\mathbf{q}}(t)) \\
    &= \mathbf{J}(\mathbf{q})^{+} \big\{\ddot{\mathbf{p}}_{des}(t) + \bm{\Lambda }_1 [\dot{\mathbf{p}}_{des}(t)- \dot{\mathbf{p}}(t)] -\mathbf{\dot{J}}(\mathbf{q})\dot{\mathbf{q}}(t) \big\}
\end{align*}
From these values, $\bm{\tau}_{in}(t)$ is defined by:
\[
    \bm{\tau}_{in}(t) = \mathbf{M}(\mathbf{q})\ddot{\mathbf{q}}_r(t) + \mathbf{C}(\mathbf{q}, \dot{\mathbf{q}})\dot{\mathbf{q}}_r(t) - \bm{\Lambda }_2 \{\dot{\mathbf{q}}(t)-\dot{\mathbf{q}}_r(t) \}
\]
In this equation, $\bm{\Lambda}_2\in\mathbb{R}^{n\times n}$
is a symmetric positive definite matrix. 

Note that this controller, suggested in \cite{nakanishi2008operational}, is equivalent to the sliding mode feedback controller introduced by \cite{slotine1987adaptive}. 
It was shown by \cite{slotine1987adaptive} that $\mathbf{p}(t)$ asymptotically converges to $\mathbf{p}_{des}(t)$.
Moreover, with this feedback controller, one can resolve kinematic singularity using a damped least-squares inverse (Section \ref{subsubsec:task_space_no_redund_discrete}) \citep{nakamura1986inverse,chiaverini1994review}.

\begin{figure*}
    \centering
    \includegraphics[trim={0.0cm 0.0cm 0.0cm 0.0cm},
     width=1.00\textwidth,clip]{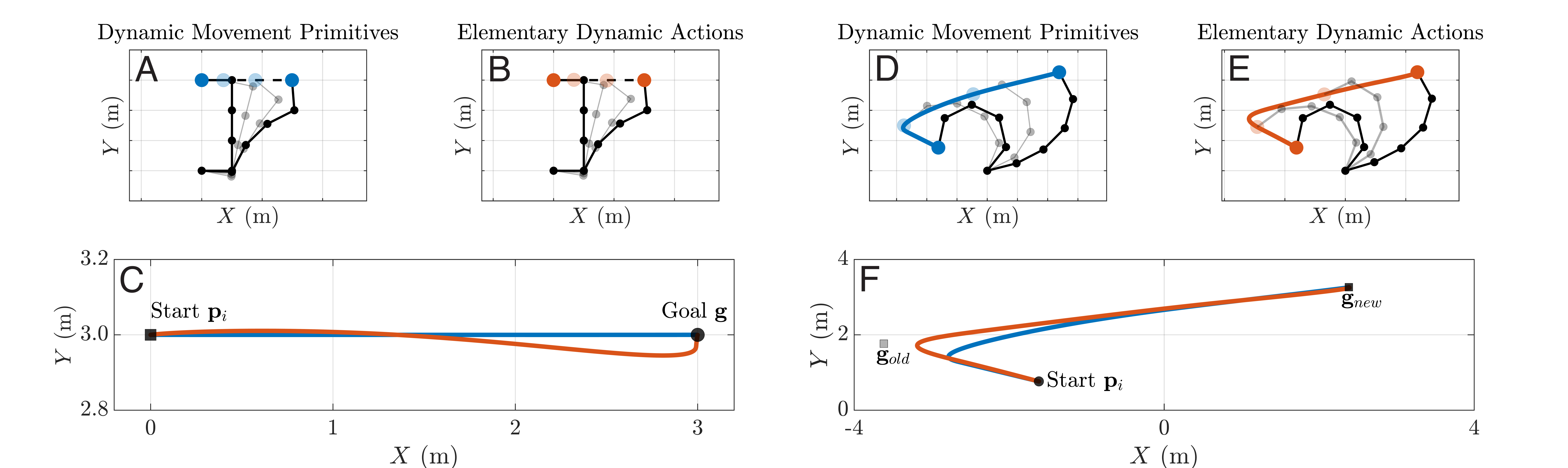}
    \caption{ Managing kinematic redundancy using Dynamic Movement Primitives (DMPs, blue) and Elementary Dynamic Actions (EDAs, orange) (Section \ref{subsubsec:managing_kinematic_redundancy}). (A, B, C) A single of discrete movement. (D, E, F) A sequence of discrete movements. The first movement headed toward $\mathbf{g}_{old}$ (square grey marker), but at time $t=t_g$, target switched location to $\mathbf{g}_{new}$ (square black marker) which necessitated a second movement.
    (A, B) The dashed black lines are a minimum-jerk trajectory. (C, F) End-effector trajectories. (A, B, C) Parameters of the minimum-jerk trajectory: $\mathbf{p}_{i}=[0,3]$m, $\mathbf{g}=[3,3]$m, $T=2.0$s. (D, E, F) Parameters of the minimum-jerk trajectories: $\mathbf{p}_{i}=[-1.62, 0.76]$m, $\mathbf{g}_{old}=[-3.62, 1.76]$m, $\mathbf{g}_{new}=[2.38, 3.26]$m, $T_1=2.0$s, $T_2=3.0$s, $t_g=1.0$. 
    Parameters of DMPs: $\bm{\Lambda}_1=80\mathbf{I}_{2}$, $\bm{\Lambda}_2=100\mathbf{I}_{5}$, $\tau=T$. Other parameters are identical to those in Figure \ref{fig:dmps_vs_edas_discrete_joint}. Parameters of EDAs: $\mathbf{K}_p=300\mathbf{I}_2$ N/m, $\mathbf{B}_p=100\mathbf{I}_2$ N$\cdot$s/m, $\mathbf{B}_q=30\mathbf{I}_5$ N$\cdot$m$\cdot$s/rad. }
    \label{fig:manage_redund}
\end{figure*}

\subsubsection{Elementary Dynamic Actions}\label{subsubsec:example_superimpose_multiple_movements}
For EDAs, kinematic redundancy of the robot manipulator can be managed by superimposing multiple mechanical impedances \citep{hermus2021exploiting,verdi2019compositional}.
In detail, a first-order task space impedance controller (Equation \eqref{eq:first_order_task_space_impedance}) can be superimposed with a joint-space controller that implements joint damping (Equation \ref{eq:joint_space_impedance_controller}):
\begin{align}\label{eq:managing_kinematic_redundancy}
    \begin{split}
        \mathbf{Z}_p(t) &= \mathbf{K}_p \{\mathbf{p}_0(t) - \mathbf{p}(t) \} + \mathbf{B}_p \{\dot{\mathbf{p}}_0(t)- \dot{\mathbf{p}}(t) \} \\
         \mathbf{Z}_q(t) &= -\mathbf{B}_q \dot{\mathbf{q}}(t) \\
        \bm{\tau}_{in}(t) &=  \mathbf{J}(\mathbf{q})^\text{T} \mathbf{Z}_p(t) + \mathbf{Z}_q(t)    
    \end{split}
\end{align}
As in Section \ref{subsubsec:joint_space_discrete_EDAs} and Section \ref{subsubsec:discrete_task_motor_primitives}, with constant symmetric positive definite matrices of $\mathbf{K}_p, \mathbf{B}_p, \mathbf{B}_q$, a goal directed discrete movement can be generated by setting $\mathbf{p}_0(t)$ to be a submovement which ends at goal location $\mathbf{g}$.

The stability of this controller is shown in \cite{arimoto2005natural, arimoto2005simple,arimoto2006human,lachner2022geometric}, where an asymptotic convergence of $\dot{\mathbf{q}}(t)\rightarrow \mathbf{0}$ and $\mathbf{p}(t)\rightarrow\mathbf{g}$ is guaranteed for discrete movement $\mathbf{p}_{0}(t)$.
This also implies an asymptotic convergence of $\mathbf{q}(t)$ to one of the infinite number of solutions that satisfies $\mathbf{g}=\mathbf{h}(\mathbf{q}(t))$.

Like the controller in Equation \eqref{eq:first_order_task_space_impedance}, this controller does not involve an inversion of the Jacobian matrix. Moreover, explicitly solving the inverse kinematics and an inverse dynamics model are not required. 
Hence, the approach remains stable near kinematic singularities.

For the desired discrete movement, we used a damping joint-space impedance operator to reduce the joint motions in the nullspace of $\mathbf{J(q)}$ (Equation \eqref{eq:managing_kinematic_redundancy}). 
Note that for repetitive rhythmic movements in task-space \citep{klein1983review}, this controller might result in non-negligible drift in joint space \citep{mussa1991integrable}. 
If this resulted in control problems, e.g., violation of joint limits, one can augment the damping impedance operator with a stiffness term $\mathbf{K}_q (\mathbf{q}_0(t) - \mathbf{q}(t))$, which will eliminate the joint-space drift and enable a stable equilibrium configuration. For brevity, this was not considered here.

Superimposing joint-space and task-space impedances can yield task conflicts, unless the virtual joint configuration to which the joint stiffness is connected is defined at the desired goal location \citep{hermus2021exploiting}. 
Often, the task conflict is resolved by using null-space projection methods, as suggested by \cite{khatib1995inertial}. 
However, it is important to note that the resultant controller violates passivity \citep{lachner2022geometric}. Alternatively, it has been shown that, with sufficiently large null-space dimension, the task conflict is minimized or even elminated (\cite{hermus2021exploiting}).

\subsubsection{Simulation Example}\label{subsubsec:managing_kinematic_redundancy}
Consider a 5-DOF planar serial-link robot model, where each link consists of a single uniform slender bar with mass and length of 1kg and 1m, respectively. 
With this robot model, we generated a single (or sequence of) discrete movement(s).
As in Section \ref{subsubsec:task_space_no_redund_discrete}, a minimum-jerk trajectory was used.
For the sequence of discrete movements, the trajectories of Section \ref{subsubsec:sequence_of_discrete_movements} were used.
The code scripts for the simulations are \texttt{main\_redundant\_discrete.py} for single discrete movement and \texttt{main\_redundant\_sequencing.py} for sequence of discrete movements.

As shown in Figure \ref{fig:manage_redund}, both approaches were able to achieve goal-directed discrete movements.
For DMPs, the approach used a feedback controller with reference trajectories generated by DMPs to manage kinematic redundancy \citep{slotine1991applied, nakanishi2008operational}. 
For EDAs, the controller simply reused the task-space impedance controller in Section \ref{subsubsec:discrete_task_motor_primitives} and combined it with the impedance controller (with $\mathbf{K}_p=\mathbf{0}$) of Section \ref{subsubsec:joint_space_discrete_EDAs}. As demonstrated in Section \ref{subsec:obstacle_avoidance}, this example again emphasizes the modular property of EDAs.

\section{Discussion}\label{discussion}
In the previous Sections, we presented detailed implementations of both DMPs and EDAs to solve eight control tasks.  
Here, we summarize the similarities and differences between the two approaches.
Moreover, we briefly discuss how these two methods might be combined to exploit the advantages of both approaches.

\subsection{ Similarities between the Two Approaches}\label{subsec:similarities_both_approaches}
DMPs and EDAs both stem from the idea of motor primitives. 
Hence, both approaches share the same principle --- using motor primitives as fundamental building blocks to parameterize a controller.
DMPs parameterize the controller with a canonical system, nonlinear forcing terms, and transformation systems (Section \ref{subsec:dynamic_movement_primitives}).
EDAs parameterize the controller with submovements, oscillations, and mechanical impedances (Section \ref{subsec:dynamic_motor_primitives})

Robot control based on motor primitives provides several advantages, and we presented eight examples.
First, by parameterizing the controller with motor primitives, the approaches provide a high level of autonomy for generating dynamic robot behavior.
Once triggered, the primitive behaviors ``play out'' without requiring intervention from higher levels of the control system.

As a result, the computational complexity of the control problem is reduced.
For instance, we showed that DMPs can be scaled to multi-DOF systems by synchronizing a canonical system with multiple transformation systems (Section \ref{subsubsec:imitation_learning}).
With Locally Weighted Regression of Imitation Learning, learning new motor skills is reduced to calculating the best-fit weights of the nonlinear forcing terms. 
The best-fit weights are learned by simple matrix algebra, which is computationally efficient (Section \ref{subsec:joint_space_trajectory_tracking}, \ref{subsec:task_space_traj_track_no_redund}).
In fact, it was reported that this computational efficiency of DMPs achieved control of a 30-DOF humanoid robot  \citep{atkeson2000using,ijspeert2002movement,schaal2007dynamics,ijspeert2013dynamical}.

For EDAs, by parameterizing the controller with motor primitives, the process of acquiring and retaining complex motor skills is simplified by identifying a reduced set of parameters, e.g., the initial and final positions of a submovement (Section \ref{subsec:goal_changing}, \ref{subsec:managing_kin_redund}).
These computational advantages are particularly prominent in control tasks associated with high-DOF systems, e.g., manipulation of flexible, high-dimensional objects \citep{nah2020dynamic,nah2021manipulating,nah2023learning}.

Moreover, motor primitives offer a modular framework for robot control.
By treating motor primitives as basic modules, acquisition or generation of new motor
skill occurs at the level of modules or their combination \citep{d2016modularity}. 
Once the modules are learned, one can generate a new repertoire of movements by simply combining or reusing the learned modules. 
As shown in several simulation examples, i.e, obstacle avoidance (Section \ref{subsec:obstacle_avoidance}), combination of discrete and rhythmic movements, (Section \ref{subsec:superposition_of_discrete_rhythmic}), sequencing discrete movements (Section \ref{subsec:goal_changing}), managing kinematic redundancy (Section \ref{subsec:managing_kin_redund}),
these tasks were simplified by the modular properties of DMPs and EDAs.
This modular property provides strong adaptability and flexibility for robot control, as learning new motor skills by combining or reusing learned modules is intuitively easier than learning ``from scratch.''
However, it is worth emphasizing that the details of the modular property are significantly different for the two approaches (Section \ref{subsubsec:discussion_modularity}). 

\subsection{ Differences between the Two Approaches}
\subsubsection{The Need for an Inverse Dynamics Model}\label{subsubsec:requirement_of_inverse_kinematics}
For torque-controlled robots, DMPs require an inverse dynamics model, whereas EDAs do not  (Section \ref{subsec:inverse_dynamics_model}).
An inverse dynamics model can introduce practical challenges and constraints when applying DMPs to torque-controlled robots. In particular, acquiring accurate models of the robot can be challenging and time-consuming.

However, the drawbacks associated with an inverse dynamics model can be dismissed for position control in joint-space. As a result, the application of DMPs in joint-space position-controlled robots is straightforward and efficient.

Nevertheless, joint-space position control presents its own challenges when compared to torque-controlled methods.
One obvious challenge is the problem of kinematic transformation between the robot's generalized coordinates  and task-space coordinates \citep{hogan2022contact}.
In fact, these challenges were  highlighted in the examples presented, e.g., the problem of inverse kinematics and kinematic singularity (Section \ref{subsec:task_space_traj_track_no_redund}) and managing kinematic redundancy (Section \ref{subsec:managing_kin_redund}).
Recall that these problems necessitate additional methods, e.g., damped least-squares inverse to manage kinematic singularity (Section \ref{subsec:task_space_traj_track_no_redund}) or sliding mode control to manage kinematic redundancy (Section \ref{subsec:managing_kin_redund}). 
Note that the eight control methods to manage kinematic redundancy presented by \cite{nakanishi2008operational} (which include the latter approach) assume feedback control using torque-actuated robots, not position-controlled robots. 

Perhaps more important, position control introduces instability for tasks involving contact and physical interaction.
Robot control involving physical interaction requires controlling the interactive dynamics between the robot and environment \citep{hogan2022contact}.
For this, using position control turns out to be inadequate \citep{de2008atlas}. 
A position-actuated robot fails to provide the level of compliance needed to achieve safe physical interaction. 
Moreover, the interactive dynamics cannot be directly regulated independent of the environment.
Instead of position control, we considered control methods using (ideal) torque-actuated robots to manage contact and physical interaction (Section \ref{subsec:modulatingMechImpedances}).

\subsubsection{Tracking Control}
In the absence of uncertainties and external disturbances, DMPs can achieve perfect trajectory tracking, both in task-space and joint-space coordinates (Section \ref{subsec:joint_space_trajectory_tracking}, \ref{subsec:task_space_traj_track_no_redund}, \ref{subsec:rhythmic_movement}, \ref{subsec:superposition_of_discrete_rhythmic}, \ref{subsec:managing_kin_redund}).
Using Imitation Learning (Section \ref{subsubsec:imitation_learning}), tracking a trajectory of arbitrary complexity can be achieved. 
DMPs also allow online trajectory modulation of the learned trajectory, which was shown in the obstacle avoidance example (Section \ref{subsec:obstacle_avoidance}). 

EDAs control neither position nor force directly (Section \ref{subsubsec:NortonEN}). 
Hence, non-negligible tracking error arises unless high values of mechanical impedances are employed (Section \ref{subsec:task_space_traj_track_no_redund}, \ref{subsec:rhythmic_movement}).
To achieve tracking control with EDAs for a given desired trajectory, an additional method to derive the corresponding virtual trajectory should be employed.
For instance, trajectory optimization methods which calculate the time course of impedances and virtual trajectories that produce the desired trajectory might be employed. 

\subsubsection{Contact and Physical Interaction}\label{subsubsec:contact_and_physical_interaction}
To guarantee robustness against contact and physical interaction, DMPs superimpose a joint-space PD controller on the feedforward torque command from the inverse dynamics model (Section \ref{subsec:inverse_dynamics_model}).
On the other hand, EDAs include mechanical impedances as a distinct class of primitives (Section \ref{subsubsec:mechanical_impedances}). 
With an appropriate choice of mechanical impedances, the approach is robust against uncertainty and unexpected physical contact \citep{hogan2022contact}. 
The dynamics of physical interaction can be directly controlled by modulating mechanical impedances (Section \ref{subsec:modulatingMechImpedances}).
By superimposing passive mechanical impedances, passivity is preserved (Section \ref{subsec:managing_kin_redund}).

Note that the equation of PD control used for DMPs (Equation \eqref{eq:dynamic_movement_primitives_superimposed}) is identical to a first-order joint-space impedance controller (Equation \eqref{eq:joint_space_impedance_controller}).
However, care is required: They are identical only if the robot actuators are ideal torque sources and the impedance is specified in joint space.
Impedance control is more general than PD control and not limited to first-order joint-space behavior; by definition, mechanical impedance determines the dynamics of physical interaction at an interaction port, which may, in principle, be at any point(s) on the robot \citep{won1997comment}. 


\subsubsection{Managing Kinematic Singularity and Redundancy}
For DMPs, control in task-space requires solving an inverse kinematics problem (Section \ref{subsec:task_space_traj_track_no_redund}).
This introduces the challenges of managing kinematic singularity and kinematic redundancy.
The latter can be resolved by using any of multiple feedback control methods presented by \citep{nakanishi2008operational}, and for the example (Section \ref{subsubsec:managing_kinematic_redundancy}) we used sliding mode control.
Nevertheless, this requires  feedback control based on an error signal.
This introduces a non-negligible error, and instead of perfect tracking, asymptotic convergence   is achieved.
Moreover, the methods still involve a Jacobian (pseudo-)inverse.
Consequently, an additional method to handle kinematic singularity should be employed.
Finally, null-space projection methods violate passivity (Section \ref{subsubsec:task_space_redund_dynamic_movement_primitives}), and advanced methods to guarantee the robot's stability might be needed \citep{dietrich2015overview,lachner2022geometric}.

For EDAs, explicitly solving the inverse kinematics is not required \citep{hogan1987stable}.
Seamless operation into and out of kinematic singularities are possible (Section \ref{subsubsec:task_space_no_redund_discrete}).
EDAs superimpose multiple mechanical impedances to manage kinemtic redundancy (Section \ref{subsubsec:managing_kinematic_redundancy}).
Unlike null-space projection methods, passivity is preserved \citep{lachner2022geometric, hogan2022contact}.

\subsubsection{Modularity in Robot Control}\label{subsubsec:discussion_modularity}
As discussed in the examples of Section \ref{subsec:comparison} and Section \ref{subsec:similarities_both_approaches}, for both approaches, motor primitives provide a modular control framework for robot control, which thereby simplifies multiple control tasks (Section \ref{subsec:obstacle_avoidance}, \ref{subsec:superposition_of_discrete_rhythmic}, \ref{subsec:goal_changing}). 
Nevertheless, it is important to note that the extent of modularity and its practical implications significantly differ between these two approaches.

A clear distinction is evident when combining multiple movements. 
For DMPs, by design, discrete and rhythmic movements are generated by different DMPs.
Hence, discrete and rhythmic movements cannot be simply superimposed (Section \ref{subsec:superposition_of_discrete_rhythmic}). 
Moreover, to sequence discrete movements, the goal location $\mathbf{g}(t)$ of the previous discrete movement is modulated (Equation \eqref{eq:DMP_goal_changing_sequence}), (Section \ref{subsec:goal_changing}).

On the other hand, for EDAs, sequencing and/or combining multiple movements can be seemlessly conducted at the level of the virtual trajectory (Section \ref{subsubsec:NortonEN}).
Recall that a combination of discrete and rhythmic movements was achieved by simply combining submovements and oscillations (Section \ref{subsec:superposition_of_discrete_rhythmic}). 
For sequencing discrete movements, the subsequent discrete movement was superimposed ``without modifying'' the previous movement (Section \ref{subsec:goal_changing}).
Moreover, one can freely choose the basis function of a submovement, rather than being restricted to a response of a stable linear system (Figure \ref{fig:discrete_and_rhythmic}).
These properties provide a notable degree of simplicity and modularity, as individual motions can be separately planned and simply superimposed without further modification.

The superposition principle of mechanical impedances enables breaking down complex tasks into simpler sub-tasks, solving each sub-task with a specific module, and simply combining these modules to solve the original problem  (Section \ref{subsec:obstacle_avoidance}).
Using mechanical impedances with this ``divide-and-conquer'' strategy, the overall complexity of the control problem can be significantly reduced. 
A task may be achieved by simply reusing the impedance controllers of component tasks without any modification.
This modular property of EDAs is in contrast with DMPs.
While the learned weights of the nonlinear forcing terms can be reused for a single DMP, multiple DMPs cannot be simply combined by merging the learned weights of different DMPs. 

\subsection{ Combining the Best of Both Approaches }\label{subsec:combine_best_of_both}
Despite these differences, both approaches may be combined to leverage their respective advantages and alleviate their limitations.
EDAs are robust against uncertainty and have advantages for physical interaction. 
DMPs can easily learn and track trajectories of arbitrary complexity.
A combination of both approaches may allow robustness against uncertainty and physical interaction, while enabling efficient learning and tracking of trajectories with arbitrary complexity.

In fact, superimposing a low-gain PD controller on the feedforward torque command of DMPs (Section \ref{subsec:modulatingMechImpedances}) can be regarded as an example of combining both approaches. 
Given an ideal torque-actuated robot, a first-order joint-space impedance controller (an example of an EDA) is added to a feedforward torque command based on DMPs. 
Robustness against uncertainty or physical contact can also be achieved by superimposing other mechanical impedances, e.g., a first-order task-space impedance controller (Equation \eqref{eq:first_order_task_space_impedance}).

A combination of both approaches may be achieved not only at the torque level, but also at the kinematic level.
Imitation Learning of DMPs can be used to generate an EDA virtual trajectory with arbitrary complexity. An example of this approach is the ``variable impedance control with policy improvement and path integral,'' introduced by \cite{buchli2011learning}.
This approach demonstrated the potential of combining DMPs and EDAs to achieve a rich set of movements that are also robust against uncertainty and physical interaction.

\section{Conclusion}
In this paper, we provided a detailed comparison of two motor-primitives approaches in robotics: DMPs and EDAs. 
Both approaches utilize motor primitives as fundamental building blocks for parameterizing a controller, enabling highly dynamic robot behavior with minimal high-level intervention. 

Despite this similarity, there are notable differences  in their implementation. 
Using simulation, we delineated the differences between DMPs and EDAs through eight robot control examples.
While DMPs can easily learn and track trajectories of arbitrary complexity, EDAs are robust against uncertainty and have advantages for physical interaction. 
Accounting for the similarities and differences of both approaches, we suggest how DMPs and EDAs can be combined to achieve a rich repertoire of movements that is also robust against uncertainty and physical interaction.

In conclusion, control approaches based on DMPs, EDAs or their combination offer valuable techniques to generate dynamic robot behavior.
By understanding their similarities and differences,  researchers may make informed decisions to select the most suitable approach for specific robot tasks and applications. 

\section*{Acknowledgements}

\subsection*{Funding}
This work was supported in part by the MIT/SUSTech Centers for Mechanical Engineering Research and Education. 
MCN was supported in part by a Mathworks Fellowship.

\subsection*{Declaration of Conflicting Interests}
The Authors declare that there is no conflict of interest.

\bibliographystyle{SageH}
\bibliography{literature}

\end{document}